\def\year{2022}\relax
\documentclass[letterpaper]{article} 
\usepackage{aaai22}  
\usepackage{times}  
\usepackage{helvet}  
\usepackage{courier}  
\usepackage[hyphens]{url}  
\usepackage{graphicx} 
\usepackage{natbib}  
\usepackage{caption} 
\DeclareCaptionStyle{ruled}{labelfont=normalfont,labelsep=colon,strut=off} 
\usepackage{algorithm}
\usepackage{algpseudocode}
\usepackage{newfloat}
\usepackage{listings}
\floatstyle{ruled}
\newfloat{listing}{tb}{lst}{}
\floatname{listing}{Listing}
\usepackage{soul}

\usepackage{thm-restate}
\usepackage{xcolor}
\usepackage{amsmath}
\usepackage{amssymb}
\usepackage{amsthm}
\usepackage{bm}
\usepackage{bbm}
\usepackage{multirow}
\usepackage{mathtools}
\usepackage{subfiles}
\newtheorem{lemma}{Lemma}
\newtheorem{theo}{Theorem}

\newtheorem{defi}{Definition}

\newtheorem{ass}{Assumption}
\newtheorem{remark}{Remark}

\DeclareMathOperator*{\argmax}{argmax}
\DeclareMathOperator*{\argmin}{argmin}

\newcommand\numberthis{\addtocounter{equation}{1}\tag{\theequation}}
\newenvironment{proof-sketch}{\noindent{\textit{Sketch of Proof.}}\hspace*{1em}}{\qed\bigskip}

\newcommand{\cheng}[1]{\textcolor{magenta}{{\bf Cheng:~}#1}}

\newcommand{\myitem}[1]{\textbf{\textcolor{blue}{(#1)}}}

\title{Gaussian Process Bandits with Aggregated Feedback}
\author{
    Mengyan Zhang, \textsuperscript{\rm 1,}\textsuperscript{\rm 2}
    Russell Tsuchida,\textsuperscript{\rm 2}
    Cheng Soon Ong \textsuperscript{\rm 1,}\textsuperscript{\rm 2}
}
\affiliations{
    \textsuperscript{\rm 1} The Australian National University 
    \textsuperscript{\rm 2} Data61, CSIRO\\
    mengyan.zhang@anu.edu.au, russell.tsuchida@data61.csiro.au, chengsoon.ong@anu.edu.au

}

\usepackage{bibentry}

\begin{document}


\maketitle

\begin{abstract}
	We consider the continuum-armed bandits problem, under a novel setting of recommending the best arms within a fixed budget under aggregated feedback. 
This is motivated by applications where the precise rewards are impossible or expensive to obtain, while an aggregated reward or feedback, such as the average over a subset, is available.
We constrain the set of reward functions by assuming that they are from a Gaussian Process and propose the Gaussian Process Optimistic Optimisation (GPOO) algorithm. 
We adaptively construct a tree with nodes as subsets of the arm space, where the feedback is the aggregated reward of representatives of a node. 
We propose a new simple regret notion with respect to aggregated feedback on the recommended arms.  
We provide theoretical analysis for the proposed algorithm, and recover single point feedback as a special case. 
We illustrate GPOO and compare it with related algorithms on simulated data.

\end{abstract}

\section{Introduction}
\label{sec:introduction}
In the continuum-armed bandit problem with a fixed budget, 
an agent adaptively chooses a sequence of $N$ options from a continuous set (\emph{arm space}) in order to minimise some objective 
given an oracle that provides noisy observations of the objective evaluated at the options~\citep{agrawal1995continuum,bubeck_conti_2011}. 
The objective may measure the total cost, for example the \emph{cumulative regret}, or may give an indication of the quality of the final choice, for example the \emph{simple regret}. 
The simple regret setting may be viewed as black-box, zeroth order optimisation of the objective under noisy observations. 
In practical settings, it is possible that one cannot observe the objective directly. This motivates a more flexible notion of an oracle. 
In this work we consider an oracle that provides noisy average evaluations of the objective over some grid (defined in a precise sense in ~\eqref{equ: ave reward defi}).

For the problem of black-box optimisation of a function $f$ under single point stochastic feedback, 
\citet{munos2014treeBandits} proposed a continuum-armed bandit algorithm called Stochastic Optimistic Optimisation (StoOO) with adaptive hierarchical partitioning of arm space,
under the \emph{optimism in the face of uncertainty principle}. 
For bandits with aggregated feedback,
\citet{rejwan2020top} studied finite-armed case for the combinatorial bandits under full-bandit feedback. 
Other related settings are discussed in~\S~\ref{sec:related_work}. 
We consider one important gap in the literature, best arm(s) identification for continuum-armed bandits with average rewards under a fixed budget.

Our goal is to recommend a local area with best average reward feedback. 
We propose \emph{aggregated regret} (Definition \ref{defi: aggregated regret}) to reflect this objective, devise an algorithm in \S~\ref{sec: GPOO}, and show upper bounds of the aggregated regret under our algorithm in \S~\ref{sec: theo}. In \S~\ref{sec: experiments}, we compare our algorithm with related algorithms in a simulated environment. 
Our algorithm shows the best empirical performance in terms of aggregated regret. 

Our \textbf{contributions} are 
(i) a new continuum-armed bandits setting under the aggregated feedback and corresponding new simple regret notion,
(ii) the first fixed budget best arms identification algorithm (GPOO) for continuum-armed bandit with noisy average feedback,
(iii) theoretical analysis for the proposed algorithm, and
(iv) empirical illustrations of the proposed algorithm.
\section{Formulation and Preliminaries}
\label{sec: formualtion and preliminaries}

\subsection{Motivation}
\begin{figure}
    \centering
    \includegraphics[scale=0.25]{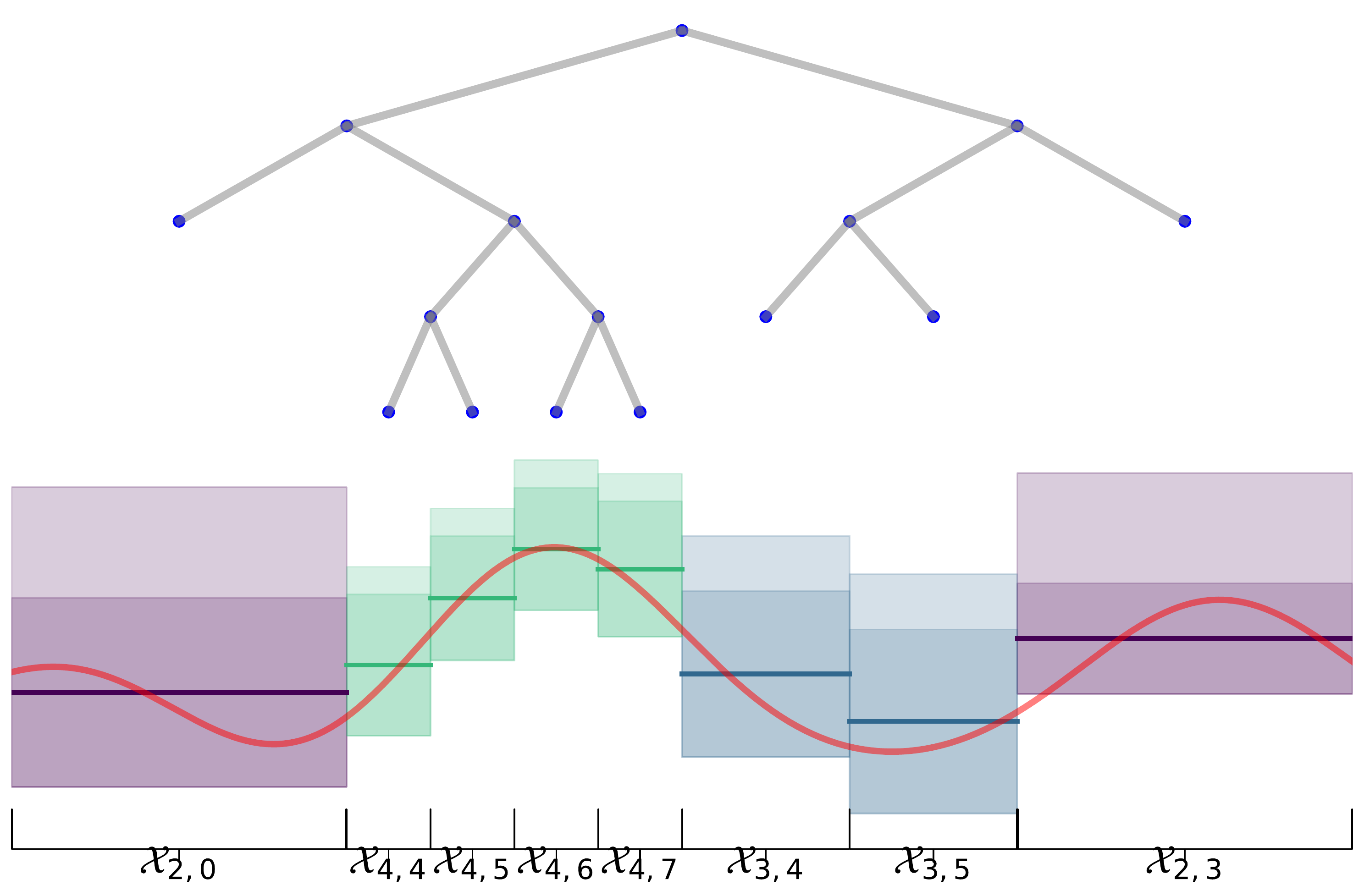}
    \caption{GPOO adaptively constructs a tree where the value associated with each node is an estimate of the aggregated reward over a cell. Red shows the reward function to be optimised. Solid horizontal lines show estimated mean aggregated reward. Dark shaded regions shows probable objective function ranges based on Bayesian uncertainty. Light shaded regions additionally account for potential function variation due to smoothness assumptions.}
    \label{fig:fig1}
\end{figure}

Two unique properties of the setting we consider are (i) the reward signal is aggregated, and (ii) the aggregation occurs on hierarchically partitioned continuous space. 
Gaussian Process Optimistic Optimisation (GPOO) makes use of both of these properties, as illustrated in Figure~\ref{fig:fig1} and described in Algorithm~\ref{alg: GPOO}.

\paragraph{Aggregated feedback.} Quantitative observations of the real-world are often made through smooth rather than instantaneous measurements. 
Average observations may arise from physical, hardware, privacy constraints.
We provide three potential applications to motivate our setting: 
\myitem{1} \textit{Radio telescope.} Arm cells are the spatial-frequency (orientation and angular resolution) coordinates of objects in the sky. The average radio wave energy (reward) can be inferred from the radio telescope for the queried area. Only the aggregated reward is observable due to frequency binning in hardware and spatial averaging.
The goal is to design a policy so that one can identify the region with the average highest radio energy with a fixed amount of querying. 
The first radio telescope that was used to detect extra-terrestrial radio sources~\citep{jansky1933electrical} and was able to determine that the source of the radiation was from the centre of the Milky Way. 
\myitem{2} \textit{Census querying.} Take the age of respondents as an example. Arms are each respondent. The oracle will return the average age (reward) of respondents inside each queried area and each query cost is the same no matter what the query is. Only the aggregated reward is allowed due to privacy concerns. 
The goal is to design a policy so that one can identify the region with the average highest age with a fixed amount of querying.
\myitem{3} \textit{DNA design}. In synthetic biology, one can modify nucleotides to control protein expression level (reward) \citep{zhang_synbio2021}. The arms are all possible DNA sequences. The goal is to find DNA sequences with highest possible protein expression level within a given budget. The experiment is expensive and the search space is too large to enumerate. We can make a mixed culture with similar DNAs in a queried feature space and measure their aggregated reward only. 


These smoothing operations present in sensor hardware designs, survey sampling methodologies and privacy-preserving data sharing motivate data analysis techniques that account for smooth or average rather than point or instantaneous measurements~\citep{zhang2020learning}. 

\paragraph{Tree structures for continuous spaces.} Function optimisation using bandits may be achieved by simultaneously estimating and maximising some estimated statistic of a black-box objective $f$. 
This usually involves an iterative algorithm, whereby at each step a point (or points) is (are) sampled and then the estimated statistic is updated and then maximised. 
The estimated statistic is called an \emph{acquisition function}, and in continuous spaces, can be computationally expensive to optimise. For example, in GP-UCB, the acquisition function is the upper confidence bound, leading to an overall computational complexity of $\mathcal{O}(N^{2d+3})$ for running the algorithm.

Hierarchically forming arms allows adaptive discretisation over the arm space, 
which provides a computationally efficient approach for exploring the continuous arm space. 
Assuming smoothness of the unknown reward function and given a budget, \citet{munos2014treeBandits} proposed a Stochastic Optimistic Optimisation (StoOO) algorithm.
They adaptively construct a tree which partitions the design space. Each leaf node in the tree represents a subset of the design space and is a candidate to be expanded. 
The expanded leaf node is chosen based on the \emph{optimisation under uncertainty criteria}. 
Here the notion of uncertainty captures both stochastic uncertainty due to reward sampling and the inherent function variation. 
The reward of the leaf node is summarised by the reward obtained by evaluating the noisy objective at a some point in the subset (\citet{munos2014treeBandits} calls these \emph{centres}, we call them \emph{representative points}).

\paragraph{Our proposed algorithm,} Gaussian Process Optimistic Optimisation (GPOO, Algorithm \ref{alg: GPOO}), extends the StoOO algorithm to the case where the $f$ is sampled from an unknown Gaussian Process (GP) and the reward feedback is an average over representatives in a subset. 
Using a GP allows us to encode smoothness assumptions on the function $f$ through a choice of kernel (see Assumptions~\ref{ass:gp_known_cov},~\ref{ass:gp_smoothness2}). It also allows us to exploit the closure of Gaussian vectors under affine maps to update our belief of $f$ under aggregated feedback in a Bayesian framework.
In order to build a well-behaved tree-structure, we have to assume a certain regularity of the tree with respect to the function $f$ (see Assumptions~\ref{ass: decreasing diameters},~\ref{ass: well-shaped cells}), mirroring those in the StoOO algorithm. Assumption~\ref{ass: decreasing diameters} ensures that as the depth of the tree grows, the allowable function variation around any node decreases. Assumption~\ref{ass: well-shaped cells} rules out nodes that represent pathologically shaped subsets of the design space, such as those consisting of sets with measure zero like single points or curves.

Our problem setting recovers the single state reward feedback as a special case. 
To the best of our knowledge, we are the first work address the continumm-armed bandits function optimisation problem under aggregated feedback. 

\subsection{Problem setting}

Let the decision space and the function to be optimised be $\mathcal{X} \subset {[0,1]}^d$ and $f:\mathcal{X} \to \mathbb{R}$ respectively. 
We consider a hierarchical partitioning of the space $\mathcal{X}$ through an adaptively-built $K$-ary tree. 
Each node $(h,i)$ in the tree is placed at a depth $h$ and an index $i$. In order to partition the space, each node $(h,i)$ is associated with an attribute $\mathcal{X}_{h,i}$ called a \emph{cell}. At depth $h$, there are $K^h$ cells. That is, $ 0 \leq i \leq K^h-1$.
For any fixed $h$, the cells form a partition of $\mathcal{X}$.
Here partition is meant in the formal sense, that is, a partition of $\mathcal{X}$ is a collection of non-empty subsets of $\mathcal{X}$ such that every $\bm{x} \in \mathcal{X}$ is in exactly one of these subsets. 

We may obtain a reward from a given node $(h,i)$ through some abstract reward signal $\mathcal{R}\big((h,i) \big),$ 
where the mapping $\mathcal{R}$ takes as input the attributes of the node $(h,i)$.
These attributes include the cell described above, and may also include other attributes like the representative points, described below. 
Here we will focus only on a special case of $\mathcal{R}$ and leave other choices of $\mathcal{R}$ for future work. 

Each node is associated with $S$ points $\bm{x}_{h,i^s}$, $1 \leq s \leq S$, where $\bm{x}_{h,i^s} \in \mathcal{X}_{h,i}$.
We stress that $S$ is a quantity associated with the problem and may \emph{not} be controlled by the agent.
We call the collection $\mathcal{C}_{h,i} = \{\bm{x}_{h,i^s}\}_{1 \leq s \leq S}$ the \emph{representative points} of $\mathcal{X}_{h,i}$ or $(h,i)$. 
The reward of each node $(h,i)$ is summarised by the average reward evaluated over the representative points of the cell.
More precisely, we denote by $X_{h, i} \in \mathbb{R}^{S \times d}$ the \emph{feature matrix} of cell $\mathcal{X}_{h,i}$ with each row being exactly one element of the representative points $\mathcal{C}_{h, i}$ (where the order of the rows is not important).
In the round $t$, we select a cell $\mathcal{X}_{h_t, i_t}$ to obtain the reward $r_t$ of cell $\mathcal{X}_{h,i}$,
\begin{align}
  \label{equ: ave reward defi}
  r_t = \bar{F}(X_{h_t, i_t}) + \epsilon_t, \quad \bar{F}(X_{h_t, i_t}) := \frac{ \sum_{\bm{x} \in \mathcal{C}_{h_t, i_t}} f(\bm{x})}{|\mathcal{C}_{h_t, i_t}|},
\end{align}
where $\epsilon_t \stackrel{i.i.d.}{\sim} \mathcal{N}(0, \sigma^2)$. 
$\mathcal{X}_N, X_N$ and $\bm{x}_N$ will respectively denote the recommended cell, feature matrix and point (if an algorithm returns these objects).
A typical goal of best arm identification with fixed budget is to minimise the simple regret \cite{audibert2010best}. 
\begin{defi}[Simple regret]
\label{defi: simple regret}
We denote an optimal arm $\bm{x}^\ast \in \argmax_{\bm{x} \in \mathcal{X}} f(\bm{x})$. Let $\bm{x}_N$ be our recommended point after $N$ rounds. The \emph{simple regret} is defined as 
  \begin{align}
      \hat{R}(\bm{x}_N) = f(\bm{x}^\ast) - f(\bm{x}_N).
  \end{align}
\end{defi}
In our setting, a slightly different surrogate notion of regret will be easier to analyse.
Correspondingly, we introduce the \emph{aggregated regret} under our setting in Definition \ref{defi: aggregated regret}. 
The goal is to minimise the aggregated regret with a fixed budget of $N$ reward evaluations.  
That is, we aim to recommend a local area with highest possible aggregated reward.
This is highly motivated by the applications where the measurement is over a local area instead of precise point querying, for example, the sensor hardware designs and radio telescope we mentioned in the introduction. 
When $S = 1$, the aggregated regret in Definition \ref{defi: aggregated regret} is the same as \emph{simple regret}.

\begin{defi}[Aggregated regret]
\label{defi: aggregated regret}
Let $\mathcal{X}_N$ be our recommended cell after $N$ rounds and $X_N$ the corresponding feature matrix. The \emph{aggregated regret} is defined as 
  \begin{align}
      R_N = f(\bm{x}^\ast) - \bar{F}(X_{N}).
  \end{align}
\end{defi}
This surrogate may be used to upper bound the simple regret. Note that $\min \limits_{\bm{x} \in \mathcal{C}_N} \hat{R}_N (\bm{x}) \leq R_N$, where $\mathcal{C}_N$ is the set of representative points in $\mathcal{X}_N$. 
Given that cell $\mathcal{X}_N$ minimises the aggregated regret, one may apply a (finite) multi-armed bandit algorithm over the set $\mathcal{C}_N$ to solve $\min \limits_{\bm{x} \in \mathcal{C}_N} \hat{R}_N (\bm{x})$.

We note that other choices of the abstract reward $\mathcal{R}$ are also natural. For example, a continuous analogue of~\eqref{equ: ave reward defi} replaces the discrete sum over $\mathcal{C}_{h_t, i_t}$ with a continuous integral over $\mathcal{X}_{h_t, i_t}$. We sketch in Appendix~\ref{app:integral_reward} how our results might extend to this setting, but leave the details for future work.

\subsection{Gaussian processes}
In order to develop our algorithm and perform our analysis, we will require $f$ to possess a degree of smoothness. We will also need to simultaneously estimate and maximise $f$.

\begin{ass}
  The black-box function $f$ is a sample from zero-mean GP with known covariance function $k$.
  \label{ass:gp_known_cov}
\end{ass}

A GP $\{ f(\bm{x}) \}_{\bm{x} \in \mathcal{X}}$ is a collection of random variables indexed by $\bm{x} \in \mathcal{X}$ such that every finite subset $\{ f(\bm{x}_i) \}_{i = 1, \ldots m}$ follows a multivariate Gaussian distribution~\citep{GPML2006}.
A GP is characterised by its mean and covariance functions, respectively
\begin{align*}
  \mu(\bm{x}) &= \mathbb{E}[f(\bm{x})] \quad \text{ and}\\
  k(\bm{x}, \bm{x}^\prime) &= \mathbb{E}[(f(\bm{x})-\mu(\bm{x}))(f(\bm{x}^\prime)-\mu(\bm{x}^\prime))].
\end{align*}
GPs find widespread use in machine learning as Bayesian functional priors. Some function of interest $f$ is \textit{a priori} believed to be drawn from a Gaussian process with some covariance function $k$ and some mean function $\mu$, where in a typical setting $\mu \equiv 0$. After observing some data, the conditional distribution of $f$ given the data, that is, the posterior of $f$, is obtained. If the likelihood is Gaussian, the prior is conjugate with the likelihood and the posterior update may be performed in closed form. We describe a precise instantiation of this update tailored to our setting in \S~\ref{sec: GPOO}.

 Modelling $f$ as a GP allows us to encode smoothness properties through an appropriate choice of covariance function and also to estimate $f$ in a Bayesian framework. This choice additionally allows us to take advantage of the closure of Gaussian distributions under affine transformations, providing us with a tool to analyse aggregated feedback.

\begin{ass}
  \label{ass:gp_smoothness2}
  The kernel $k$ 
  is such that for all $j = 1, \dots, d$,  some $a, b > 0$ and any $L>0$,
  \begin{align*}
      \mathbb{P}\left( \sup_{ \bm{x}\in D} |\partial f/\partial x_j| \geq L\right) &\leq a \exp \left(-\frac{L^2 b}{2 }\right).
  \end{align*}
\end{ass}
Assumption~\ref{ass:gp_smoothness2} implies a tail bound on $|f(\bm{x}_1) - f(\bm{x}_2)|$, and may be shown to hold for a wide class of covariance functions including the squared exponential and Mat\'ern class with smoothness $\nu > 2$. Let $\ell$ denote the the $L_1$ distance. By directly applying Theorem 5 of~\citet{ghosal2006posterior}, we show the following in Appendix~\ref{app:gp_tail}.

\begin{restatable}{prop}{Smooth}
\label{prop: high prop smoothness}
  Assumption~\ref{ass:gp_smoothness2} implies that for some constants $a, b > 0$ and any $L>0$, $\bm{x}_1, \bm{x}_2 \in \mathcal{X}$,
      \begin{align*}
      \mathbb{P}\big( |f(\bm{x}_1) - f(\bm{x}_2)| \geq L \ell(\bm{x}_1, \bm{x}_2) \big) \leq  a e^{-L^2 b / 2}. 
  \end{align*}
  If $\sup \limits_{x \in \mathcal{X}} \frac{\partial^2}{\partial x_j \partial x'_j} k(x, x') \big| _{x=x'} < \infty$ and $k$ has mixed derivatives of order at least $4$, then $k$ satisfies Assumption~\ref{ass:gp_smoothness2}.
\end{restatable}

Assumptions~\ref{ass:gp_known_cov} and~\ref{ass:gp_smoothness2} were also made by~\cite{srinivas2009_GPUCB}, but to the best of our knowledge we are the first to exploit closure of GPs under affine maps in the setting of best arm identification under fixed budget.

\section{Algorithm: GPOO}
\label{sec: GPOO}
Inspired by StoOO \cite{munos2014treeBandits}, we propose Gaussian Process Optimistic Optimisation (GPOO), under the formulations introduced in \S~\ref{sec: formualtion and preliminaries}. In order to describe our algorithm, we first describe how to compute the posterior predictive GP.
\subsection{Gaussian process posterior update}
In round $t \in {1, \dots, N}$, we represent our prior belief over $f$ using a GP. Our prior in round $t$ is the posterior after so-far observed noisy observations of groups up to round $t$. The posterior update is a minor adjustment to the typical posterior inference step that would be employed if we were to consider only single point (not aggregated) reward feedback, exploiting the fact that multivariate Gaussian vectors are closed under affine transformations.

Define $X_{1:t} \in \mathbb{R}^{tS \times d}$ to be the vertical concatenation of $X_{h_j, i_j}$ for $j =1, \ldots, t$.
Similarly define $Y_{1:t} \in \mathbb{R}^t$ to be a vector with $j$th row equal to $r_j$, where $j = 1, \dots, t$.
In round $t$ we observe the feature-reward tuple $\Big(X_{h_t,i_t}, r_t \Big)$. We may write $r_t = \bm{a}^\top f(X_{h_t, i_t})+ \epsilon_t$ for corresponding $\bm{a} \in \mathbb{R}^{S \times 1}$ having all entries equal to $1/S$.

More compactly, all rounds $t \in \{1, \ldots, N\}$ may be represented through a vector equation. Let $A_{1:t} \in \mathbb{R}^{t \times tS}$ with $pq$th entry equal to $1/K$ if the $q$th element of $X_{1:t}$ is sampled at round $p$ and zero otherwise. 
In what follows,  we define $$Z_{1:t}:=(X_{1:t}, Y_{1:t}, A_{1:t}),$$ which completely characterises the history of observations up to round $t$. 
We may write $Y_{1:t} = A_{1:t} f(X_{1:t}) + \epsilon_{1:t}$, where $\epsilon_{1:t} \in \mathbb{R}^{t}$ denotes a vector with $i$th entry $\epsilon_i$.

Let $X_\ast \in \mathbb{R}^{n_\ast \times d}$ denote a matrix of test indices, and let $\bm{a}_\ast \in \mathbb{R}^{n_\ast \times 1}$ denote some corresponding weights. The posterior predictive distribution of $\bm{a}_\ast^\top f(X_\ast)$ given all the history $Z_{1:t}$ up to round $t$ is also a Gaussian and satisfies
\begin{align*}
    \bm{a}_\ast^\top f(X_\ast) \mid Z_{1:t} &\sim \mathcal{N}\left( \bm{a}_\ast^\top \mu(X_\ast\mid Z_{1:t}), \bm{a}_\ast^\top \Sigma(X_\ast \mid Z_{1:t}) \bm{a}_\ast\right), \\
    \mu(X_\ast\mid Z_{1:t}) &= M Y_{1:t}, \\
    \Sigma(X_\ast \mid Z_{1:t}) &= k\left(X_\ast, X_\ast \right)
    - M A_{1:t} k(X_t, X_\ast), \numberthis \label{equ:gp_posterior}
\end{align*}
where
\begin{align*}
    M = k(X_t, X_\ast)^{T} {A_{1:t}}^\top 
    \left( A_{1:t} k(X_t, X_t) {A_{1:t}}^\top
    +\sigma^{2} \boldsymbol{I}\right)^{-1}.
\end{align*}
With an iterative Cholesky update we may perform all $N$ inference steps in one $\mathcal{O}(N^3)$ sweep. See Appendix~\ref{app:cholesky_update}.

\subsection{Notions of uncertainty and function variation}
\begin{figure}[t!]
  \centering
  \includegraphics[scale = 0.23]{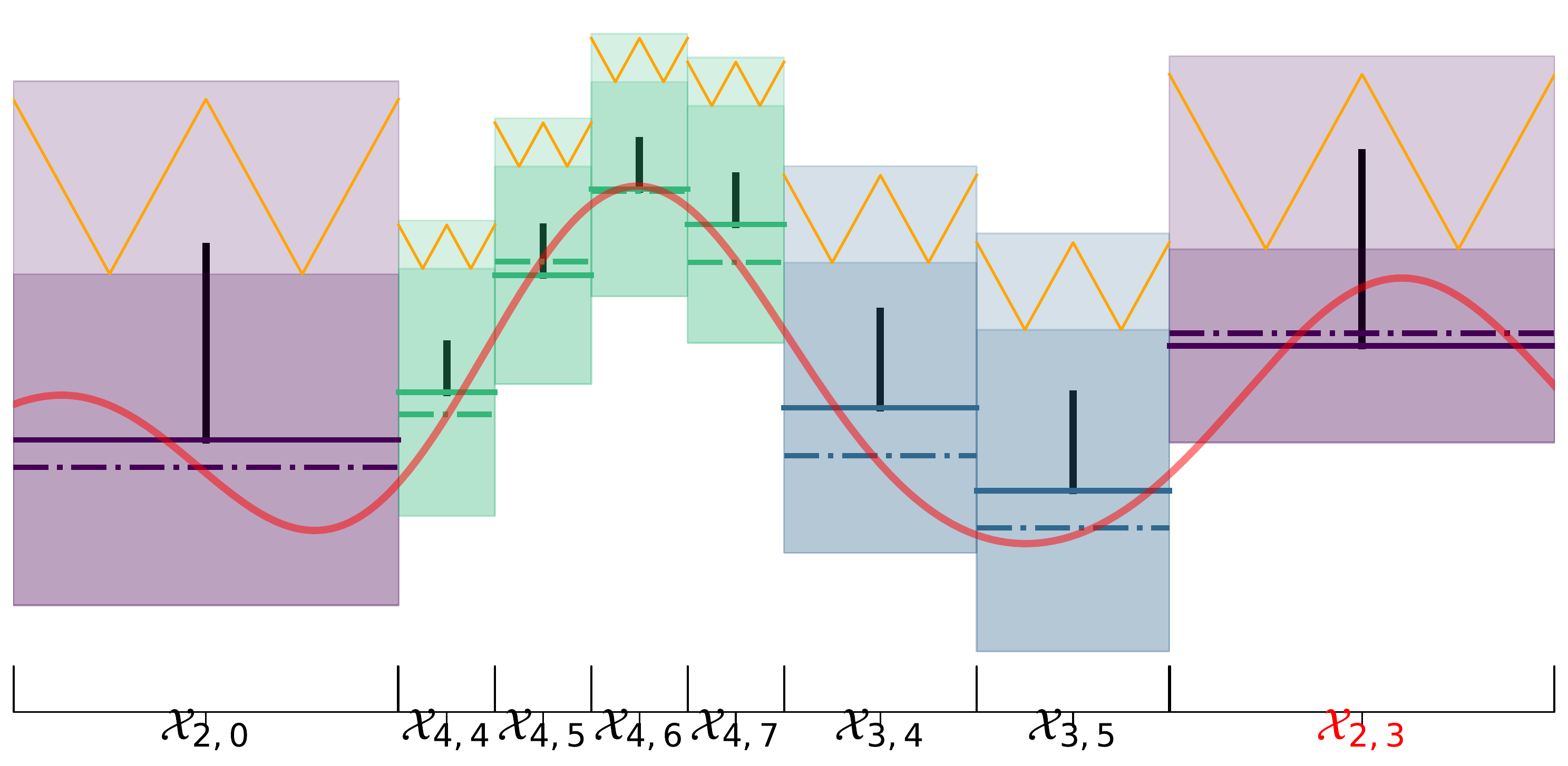}
  \includegraphics[scale = 0.23]{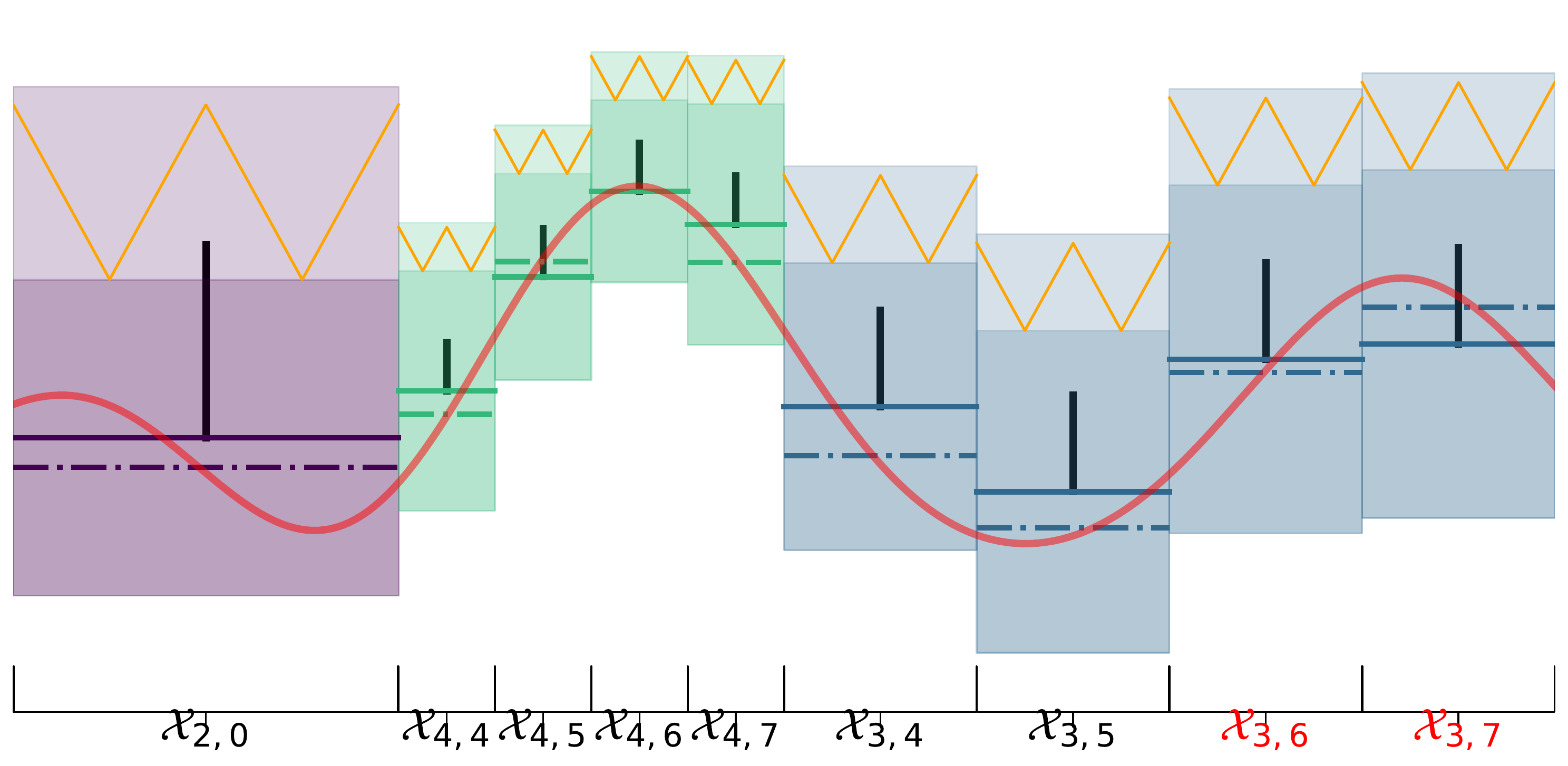}
  \caption{Quantities taken just before line $4$ of Algorithm~\ref{alg: GPOO} for $t=9$ (top) and $t=10$ (bottom). For each cell $\mathcal{X}_{h,i}$, solid and dashed horizontal lines show ${\bm{a}^\top \mu(X_{h,i} \mid Z_{t})}$ and ${\overline{F}(X_{h,i})}$ respectively. Dark shaded regions show ${CI_t(X_{h,i})}$ with colours indicating depth of cell. One-sided $b$-values are also shown by light shaded regions. Vertical black bars indicate $\delta(h)$. Shown in orange is a probable Lipschitz bound on the function value, due to event $\xi$ and Proposition~\ref{prop: high prop smoothness}. Here $S=2$ and the representative points form a uniform grid, so that the Lipschitz bounds form a W shape.
  This bound is in turn bounded by the $b$-values. When $t=9$, $\mathcal{X}_{2,3}$ is expanded since it has the highest $b$-value. When $t=10$, no cell will be expanded since $\delta(4) < CI_{10}(X_{4,6})$; instead the estimate of the function will be refined by sampling the reward of $\mathcal{X}_{4,6}$.}
  \label{fig:gpoo_quantities}
\end{figure}

We introduce the key concept for selecting which node to sample, the \emph{$b$-value} $b\big( X_\ast \mid \beta, Z, \bm{a}_\ast \big)$, which is the sum of three terms: 
the posterior mean of the corresponding feature matrix $X_\ast$, 
the confidence interval and function variation, 
\begin{align*}
  &b\big( X_\ast \mid \beta, Z, \bm{a}_\ast \big) :=\\
  &\qquad \bm{a}_\ast^\top \mu(X_\ast \mid Z) + CI(X_\ast \mid \beta, Z, \bm{a}_\ast) + \delta(h).
\end{align*}
The confidence interval is defined in terms of the posterior variance and an exploitation/exploration parameter $\beta$,
\begin{align*}
    CI\big( X_\ast \mid \beta, Z, \bm{a}_\ast \big) :&= \beta^{1/2} \sqrt{\bm{a}_\ast^\top \Sigma(X \mid Z) \bm{a}_\ast}.
\end{align*}
The last term $\delta(h)$ is the smoothness function depending on the node depth $h$ 
that satisfies Assumption~\ref{ass: decreasing diameters}.
$\delta(h)$ gives an upper bound of the function deviation of a cell in a given depth.
Figure \ref{fig:gpoo_quantities} illustrates the above quantities. 
We further define time-dependent $b$-value and confidence interval
\begin{align}
b_{h, i}(t) &:= b\big( X_{h_t, i_t} \mid \beta_t, Z_{1:t-1}, \bm{a} \big) \quad \text{ and} \numberthis \label{equ: b-value of GPOO} \\
    {CI}_t(X_\ast) &:= CI\big( X_\ast \mid \beta_t, Z_{1:t-1}, \bm{a} \big), \numberthis \label{equ: confidence interval of GPOO}
\end{align} 
where $\beta_{t} \sim \mathcal{O}(\log t)$ and will be specified in Lemma \ref{lemma: tail bound on event xi}.

\begin{ass}[Decreasing diameter]
  \label{ass: decreasing diameters}
  There exists a decreasing sequence $\delta(h) > 0$ such that for some ${L>0}$ (in Assumption \ref{ass:gp_smoothness2}), any depth $h$ and any cell $\mathcal{X}_{h,i}$, 
  $\sup _{\bm{x} \in \mathcal{X}_{h, i}} L \ell \left(\bm{x}_{h, i}, \bm{x}\right) \leq \delta(h)$ for any representative point $\bm{x}_{h,i}$.
\end{ass}

Assumption~\ref{ass: decreasing diameters} means that at any depth $h$, the largest possible distance $\sup _{\bm{x} \in \mathcal{X}_{h, i}} L \ell \left(\bm{x}_{h, i}, \bm{x}\right)$ between any point to any representative point within cell $\mathcal{X}_{h, i}$ is decreasing with respect to $h$. 
This is intuitive since we have assumed smoothness (Assumption \ref{ass:gp_smoothness2}) and constructed cells hierarchically. 
This assumption links the distance in reward space to a $\delta(h)$, which is the core concept in theoretical analysis (see Figure \ref{fig: proof roadmap}).
Compared with \citet{munos2014treeBandits}, we introduce a smoothness parameter $L$ to suit our analysis with GPs.

\begin{algorithm}[t!]
  \caption{GPOO}
  \label{alg: GPOO}
  \textbf{Input:} natural number K (-ary tree), function $f$ to be optimised, smoothness function $\delta$, budget $N$, the maximum depth nodes can be expanded $h_{\max}$.\\
    \textbf{Init:} tree $\mathcal{T}_0 = \{(0,0)\}$ (corresponds to $\mathcal{X}$), 
    leaves $\mathcal{L}_0 = \mathcal{T}_0$.
  \begin{algorithmic}[1]
    \For{$t = 1$ to $N$}
      \State{Select any ${(h_t, i_t) \in \argmax_{(h, i) \in \mathcal{L}_{t-1}} 
      b_{h,i}(t)}$.}
      \State{Observe reward $r_t = \overline{F}(X_{h_t, i_t}) + \epsilon_t$.
      }
      \State{Update posteriors $\bm{a}^\top \mu(\cdot|Z_{1:t}), \bm{a}^\top \Sigma(\cdot|Z_{1:t}) \bm{a}$} 
      \State{Update confidence intervals and $b$-values
      for all \par nodes $(h,i) \in \mathcal{L}_{t-1}$ according to E.q. (\ref{equ: confidence interval of GPOO})(\ref{equ: b-value of GPOO}).}
      \If{$\delta(h_t) \geq CI_t(X_{h_t, i_t})$ and $h_t \leq h_{\max}$}
        \State{Expand node $(h_t, i_t)$ (partition $\mathcal{X}_{h_t, i_t}$ into $K$ \par subsets) into children nodes \par ${\mathcal{C}_t = \{(h_t+1, i_1), \dots, (h_t + 1, i_K)\}}$.
        \par $\mathcal{T}_t = \mathcal{T}_{t-1} \cup \mathcal{C}_t$.}
        \State{$\mathcal{L}_{t-1} = \mathcal{L}_{t-1} \backslash \{(h_t, i_t)\}; \mathcal{L}_t = \mathcal{L}_{t-1} \cup \mathcal{C}_t$.}
      \EndIf
    \EndFor
    \State{\textbf{Return} The node with index $(h^\prime,i^\prime)$ such that $h^\prime = \argmax_{h| (h,i) \in \mathcal{T}_N \backslash \mathcal{L}_N} h.$ and ${i^\prime = \argmax_{i|(h^\prime, i) \in \mathcal{T}_N} \mu(X_{h^\prime, i}|Z_{1:t})}$}
  \end{algorithmic}
\end{algorithm}

We present the GPOO in Algorithm~\ref{alg: GPOO}. 
The key idea is that we construct a tree by adaptively discretising over the arm space.
The algorithm includes two main parts:\\
\textbf{Select node and update:}
In each round $t \in [1,N]$, a node $(h_t, i_t)$ is selected from leaves with the highest $b$-value~\eqref{equ: b-value of GPOO}.
The reward is sampled as the average of the function value over $S$ representative points plus Gaussian noise.
We then update the posteriors over cells of leaf nodes, allowing the confidence interval and $b$-values to also be updated. \\
\textbf{Expand node:}
As we increase the number of samples for a given node, the confidence interval of that node continues to decrease.
When the confidence interval is smaller than or equal to the function variation, our function estimate is more precise than the current cell range.
That is, when the condition on line 6 satisfied, the node can be expanded into $K$ children by partitioning cells into $K$ subsets. 
When the budget is reached, we return the node with the highest posterior mean prediction among non-leaf nodes. 
We select recommendations from non-leaf nodes because predictions of non-leaf nodes (satisfying event $\xi_2$ in Section \ref{sec: theo}) are more precise than leaves.
Figure~\ref{fig:gpoo_quantities} shows various quantities in the algorithm over two iterations.

\begin{remark}[Comparison to StoOO]
  (i) StoOO assumes known smoothness of the function but does not utilise correlations between arms and recommend based on predictions. GPOO recommends nodes with the GP predictions. 
  (ii) GPOO can address aggregated feedback, while StoOO can only deal with single state feedback. We also extend StoOO to address aggregated feedback in Appendix \ref{sec: AVE-StoOO}. 
\end{remark}

\begin{remark}[Computational Complexity]
The cost of performing global maximisation of an acquisition function can be exponential in the design space dimension $d$. 
The acquisition function for GP-UCB is the UCB at each point $\bm{x} \in \mathcal{X}$, leading to a total computational cost of $\mathcal{O}(N^{2d+3})$.
The problem of selecting an optimal arm from a finite subset of arms is a case of combinatorial optimisation, and the computational cost can be exponential in the number of arms.
For GPOO, for every round $t$, we consume $\mathcal{O}(t^2)$ for the GP inference procedure. Every time a leaf node is expanded, $K-1$ new leaf nodes are added to the tree, so that less than $(K-1)t$ $b$-values and confidence intervals must be computed. This leads to a total cost of $\mathcal{O}(N^4 (K-1))$.
\end{remark}

\section{Theoretical Analysis for Aggregated Regret}
\label{sec: theo}

\begin{figure}[t!]
  \centering
  \includegraphics[scale = 0.35]{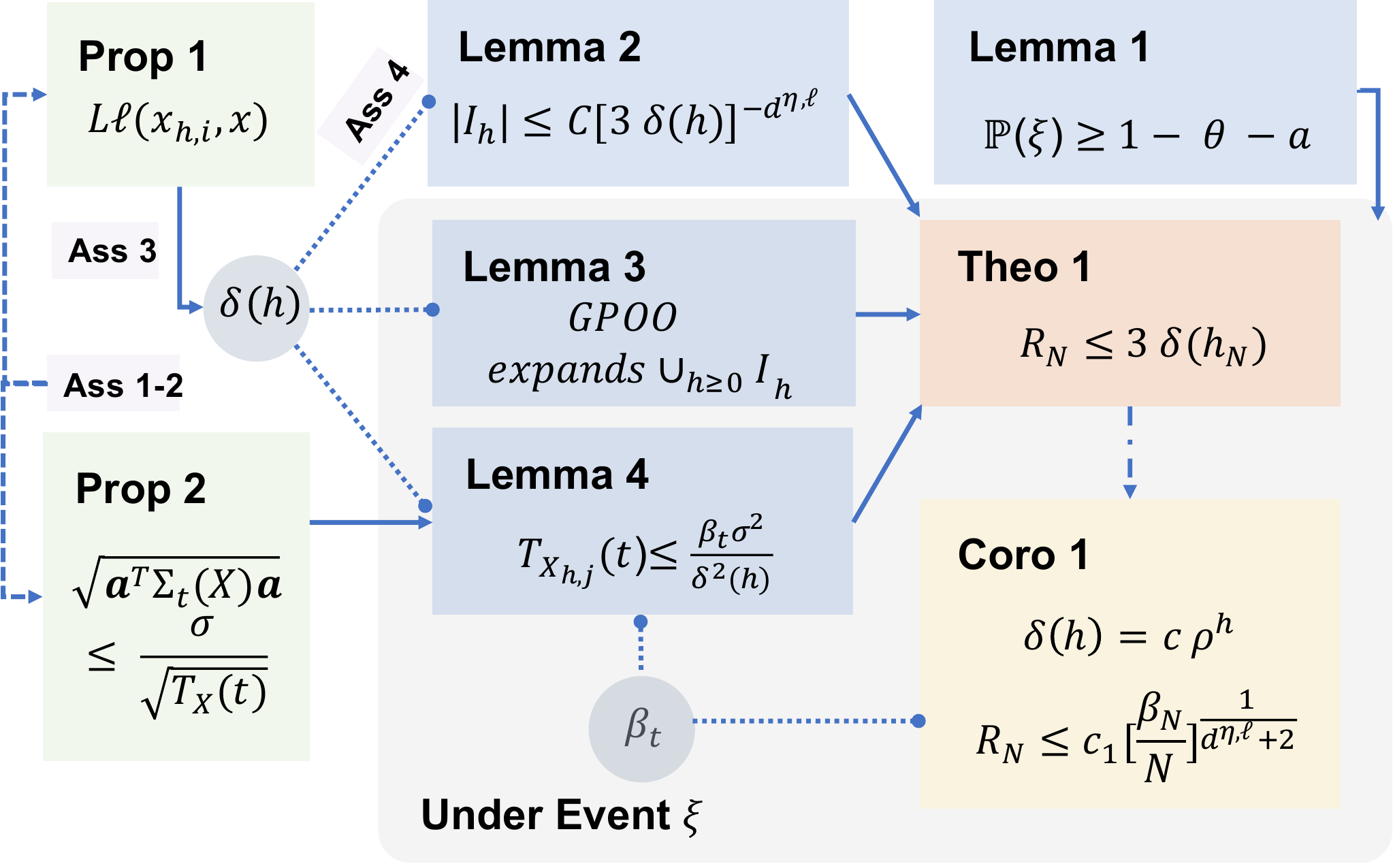}
  \caption{Proof Roadmap}
  \label{fig: proof roadmap}
\end{figure}


In this section, we provide the theoretical analysis of GPOO.
We show our proof roadmap in Figure \ref{fig: proof roadmap}.
The full proofs can be found in Appendix \ref{sec: GPOO proofs appendix}.
Our technical contribution is adapting the analysis of the hierarchical partition idea \citep{munos2014treeBandits} to the case reward is sampled from a GP \citep{srinivas2009_GPUCB} and is aggregated. 

Recall $N$ is the budget, $h_{\max}$ is a parameter representing the deepest allowable depth of tree, $\bm{x}^\ast \in \argmax \limits_{\bm{x} \in \mathcal{X}} f(\bm{x})$ is an optimal point in input space and $f^\ast = f(\bm{x}^\ast)$.
Let $\bm{x}_{h_{t}^{*}, j_{t}^{*}}$ be a representative point inside the cell $\mathcal{X}_{h_{t}^{*},j_t^{*}}$ containing $\bm{x}^\ast$ in round $t$.
We first define event $\xi$, under which we present our aggregated regret upper bound in Theorem \ref{theo: regret bound}.
Define event $\xi$ as $\xi = \xi_1 \cap \xi_2$, where
\begin{align*}
    \xi_1 :=& \left\{\forall 1 \leq t \leq N  \quad f^{*} - f\left(\bm{x}_{h_{t}^{*}, j_{t}^{*}}\right) \leq L \ell(\bm{x}_{h_{t}^{*}, j_{t}^{*}}, \bm{x}^\ast)\right\},\\
    \xi_2 :=& \big\{\forall 0 \leq h \leq h_{\max}, 0 \leq i<K^{h}, 1 \leq t \leq N \\
    & \qquad \big|\bm{a}^\top \mu(X_{h, i} \mid Z_{t-1}) -\overline{F}(X_{h, i})\big| \leq CI_t(X_{h,i} )\big\}.
\end{align*}

The event $\xi$ provides the probabilities environment for our theoretical analysis, which is the union of two events:
event $\xi_1$ describes the function $f$ local smoothness around its maximum;
event $\xi_2$ captures a concentration property from the estimation to representative summary statistics of reward,
which has been shown as an critical part for the regret analysis \cite{lattimore2018bandit, zhang2021quantile}.
In the following lemma, $b$ is the constant in Proposition~\ref{prop: high prop smoothness}.
Since $\theta$ is positive and less than $1-a$, $\mathbb{P}(\xi) \geqslant 0$.
\begin{restatable}{lemma}{xiEvent}
    \label{lemma: tail bound on event xi}
    Define $a = h_{\max} \exp(- \frac{L^2 b}{2})$
    with $L$ as a constant specified in Assumption \ref{ass: decreasing diameters}. 
    Let $\beta_{t}= 2 \log \left(M \pi_{t} / \theta\right)$,
    where $M = \sum_{h=0}^{h_{\max}} K^h$, 
    $\pi_{t}=\pi^{2} t^{2} / 6$.
    For $K$-ary tree and all $\theta \in (0, 1 - a)$,
    we have
    $\mathbb{P}(\xi) \geq 1 - a - \theta$.
\end{restatable}



To obtain the regret bound for our algorithm, we need to upper bound two pieces: the number of expanded nodes in each depth of the tree (Lemma \ref{lemma: upper bound of Ih GPOO}), 
which depends on the concept of near-optimality dimension in Definition \ref{defi: near-optimality dimension} proposed in \citet{munos2014treeBandits};
and the number of times a node can be sampled before expansion (Lemma \ref{lemma: upper bound for number of draws average}), which can be inferred from the GP posterior variance upper bound  (Proposition \ref{prop:gp_variance_bound}).
We define the set of expanded nodes at depth $h$ as $I_h$ 
\begin{align}
\label{equ: Ih}
    I_h:= \{(h,i)| \bar{F}\left(X_{h_t, j_t}\right) + 3 \delta(h) \geq f^\ast\}.
\end{align}

To upper bound the number of nodes in $I_h$, we introduce the concept of near-optimality dimension in Definition \ref{defi: near-optimality dimension}, which relates to function $f$, $\ell$ and depends on the constant $\eta$.

\begin{defi}[$\epsilon-$optimal state]
  \label{defi: epsilon optimal state}
  For any $\epsilon > 0$, define the $\epsilon-$optimal states as 
    $\mathcal{X}_\epsilon := \{\bm{x} \in \mathcal{X}| f(\bm{x}) \geq f^\ast - \epsilon\}.$
\end{defi}

\begin{defi}[$(\eta, \ell)$-near-optimality dimension $d^{\eta,\ell}$, \cite{munos2014treeBandits}]
  \label{defi: near-optimality dimension}
  For any $\epsilon > 0$, with $\epsilon-$optimal states $\mathcal{X}_\epsilon$ defined in Defition \ref{defi: epsilon optimal state},
  $d^{\eta,\ell}$  is the smallest $d \geq 0$ such that there exists $C > 0$ such that
  the maximal number of disjoint $\ell-$balls of radius $\eta \epsilon$ with centre in $\mathcal{X}_\epsilon$ is less than $C \epsilon^{-d}$.
  
  
\end{defi}

We further introduce the well-shaped cells assumption, which implies that the cells partitioned by our algorithm should contain each representatives points in a $\ell-$ball, e.g. the representatives points should not be on the boundary. 
This helps us upper bound the number of expanded nodes (Lemma \ref{lemma: upper bound of Ih GPOO}).
Unlike \citet{munos2014treeBandits}, we require it to hold for any representative point to suit our aggregated feedback setting. 
We then show that only the set of nodes in $I_h$ are expanded with GPOO in Lemma \ref{lemma：GPOO expanded nodes Ih}.

\begin{ass}[Well-shaped cells]
  \label{ass: well-shaped cells}
  There exists $\nu > 0$ s.t. for any depth $h \geq 0$, any cell $\mathcal{X}_{h,i}$ contains an $\ell$-ball 
  of radius $\nu \delta(h)$ centred in each point in the representative set $\bm{x} \in \mathcal{C}_{h,i}$.
\end{ass}

\begin{restatable}{lemma}{IhBound}
  \label{lemma: upper bound of Ih GPOO}
    Under Assumption \ref{ass: well-shaped cells}, 
      $|I_h| \leq C [3 \delta(h)]^{-d^{\eta, \ell}}.$
\end{restatable}


\begin{restatable}{lemma}{Ih}
  \label{lemma：GPOO expanded nodes Ih}
  Under event $\xi$ and Assumption \ref{ass: decreasing diameters} , GPOO only expands nodes in the set $I: \cup_{h \geq 0} I_h$.
\end{restatable}


  

We now move to derive an upper bound of the number of draws of expanded nodes. 
Following Proposition $3$ of~\citet{shekhar2018GPBanditDiscre}, using mutual information, 
we upper bound the GP posterior variance in Proposition \ref{prop:gp_variance_bound}.

\begin{restatable}{prop}{GPVarBound}
  \label{prop:gp_variance_bound}
  Let $f$ be a sample from a GP with zero mean and covariance function $k$. Let $X \in \mathbb{R}^{S \times d}$ and let $\Sigma_t(X)$ denote the posterior predictive GP covariance $\Sigma(X \mid Z_{1:t})$ according to~\eqref{equ:gp_posterior}. 
  If the history $Z_{1:t}$ contains at least $T_X(t)$ observations following the model $Y=\bm{a}^\top f(X) + \epsilon$ where $\epsilon \sim \mathcal{N}(0, \sigma^2)$,
  then we have 
      $\sqrt{ \bm{a}^\top \Sigma_t(X) \bm{a}} \leq \frac{\sigma}{\sqrt{T_X(t)}}$,
  where $\bm{a} \in \mathbb{R}^{S \times 1}$ has all entries equal to $1/S$.
\end{restatable}

With Proposition \ref{prop:gp_variance_bound}, we upper bound the number of draws for any nodes under event $\xi$ in the following lemma.

\begin{restatable}{lemma}{DrawBound}
\label{lemma: upper bound for number of draws average}
  Under event $\xi$, suppose a node $(h,j)$ (with corresponding feature matrix $X_{h,j}$ as defined in \S~\ref{sec: formualtion and preliminaries}) is sampled at least $T_{X_{h,j}}(t)$ times up to round $t$.
  Then we have
      $T_{X_{h,j}}(t) \leq \frac{\beta_{t} \sigma^2}{\delta^2(h)}$,
  where $\sigma^2$ is the noise variance, $\beta_t$ is defined in Lemma \ref{lemma: tail bound on event xi}. 
\end{restatable}

We show the aggregated regret bound for any $\delta$ under our assumptions in Theorem \ref{theo: regret bound},
and show a special case of exponential diameter in Corollary~\ref{coro: Regret bound for exponential diameters}.
  
\begin{restatable}{theo}{RegretBound}
  \label{theo: regret bound}
Define $h_N$ as the smallest integer $h^\prime$ up to round $N$ such that 
\begin{align}
\label{equ: GPOO loss bound hn defi ave}
    \frac{N}{\beta_{N}} \leq K \sum_{h = 0}^{h'} C [3 \delta(h)]^{-d^{\eta, \ell}} \frac{\sigma^2}{\delta^2(h)}.
\end{align}

For constant $a$ specified in Lemma \ref{lemma: tail bound on event xi}, and $\theta \in (0, 1 - a)$, with probability $1-\theta-a$, the simple regret of GPOO satisfies
\begin{align*}
    R_N \leq 3 \delta(h_N).
\end{align*}
\end{restatable}


\begin{restatable}[Regret bound for exponential diameters]{coro}{RegretCoro}
  \label{coro: Regret bound for exponential diameters}
  Assume $\delta(h) = c \rho^h$ for some constants $c > 0$, $\rho<1$.
  For constant $a$ specified in Lemma \ref{lemma: tail bound on event xi}, and $\theta \in (0, 1 - a)$, with probability $1-\theta-a$, the simple regret of GPOO satisfies
  \begin{align*}
      R_N \leq c_1 \Big[\frac{\beta_N}{N}\Big]^{\frac{1}{d^{\eta, \ell} + 2}},
  \end{align*}
  where $c_1 = \frac{3^{-d^{\eta, \ell}}  K C \sigma^2 }{\rho^{-(d^{\eta, \ell} + 2)} - 1}$.
  Recall $\beta_N$ is in rate $\mathcal{O}(\log N)$.
\end{restatable}

\section{Experiments}
\label{sec: experiments}


We investigate the empirical performance of GPOO on simulated data. 
We compare the aggregated regret obtained by GPOO with related algorithms,
and illustrate how different parameters influence performance.


\begin{figure}[t!]
    \centering
    \includegraphics[scale=0.25]{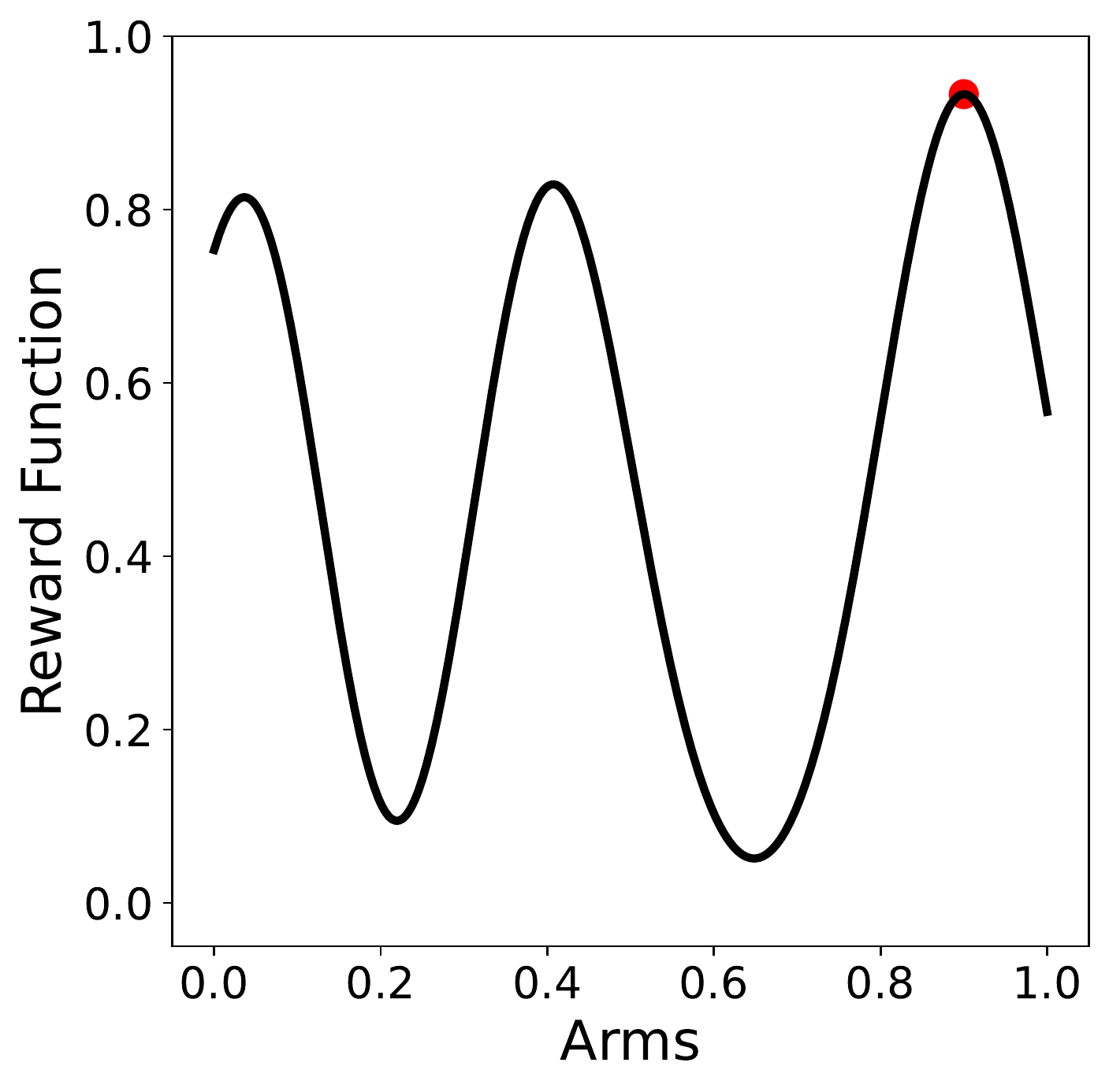}
    \includegraphics[scale=0.25]{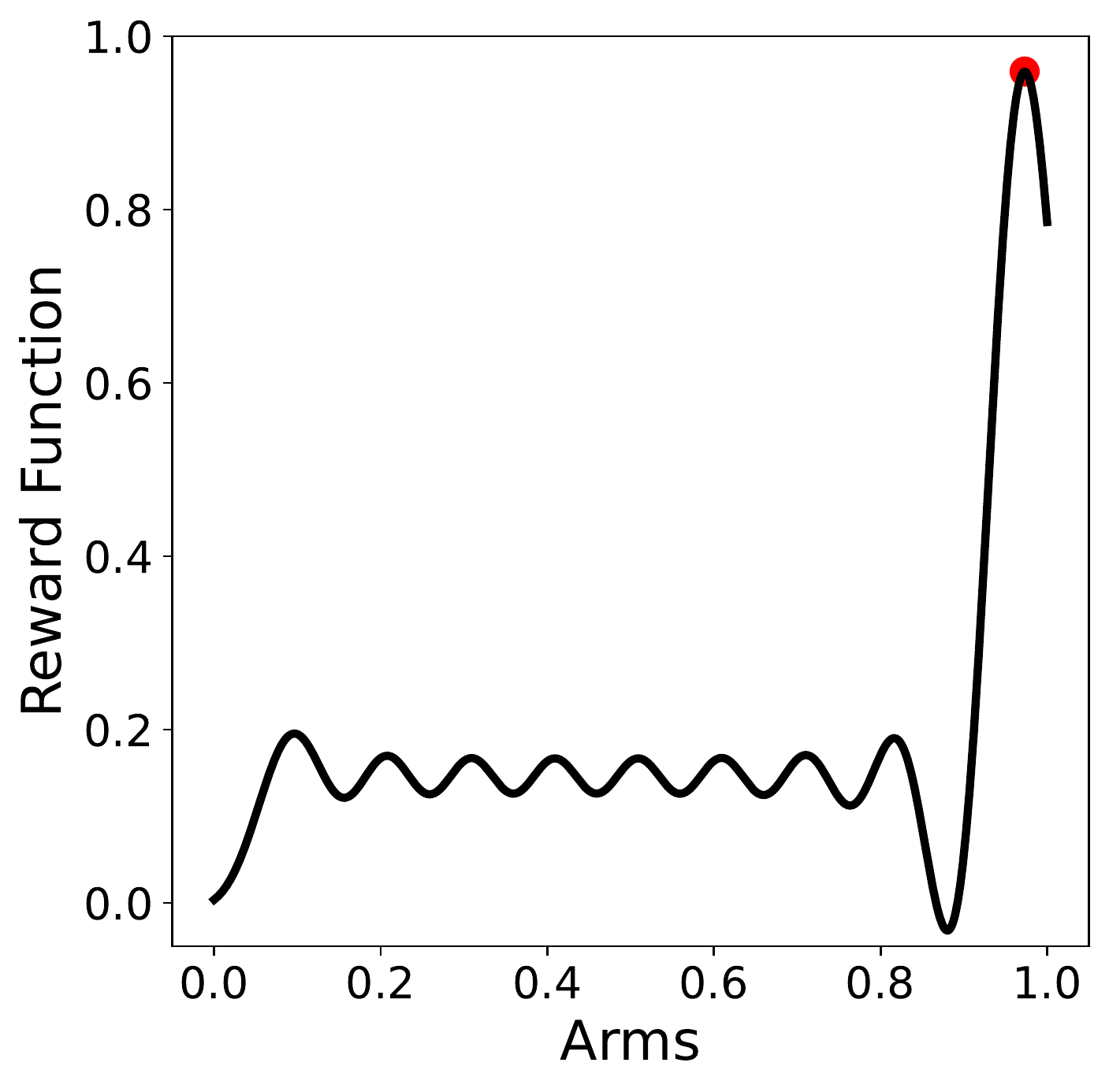}
    \includegraphics[scale=0.25]{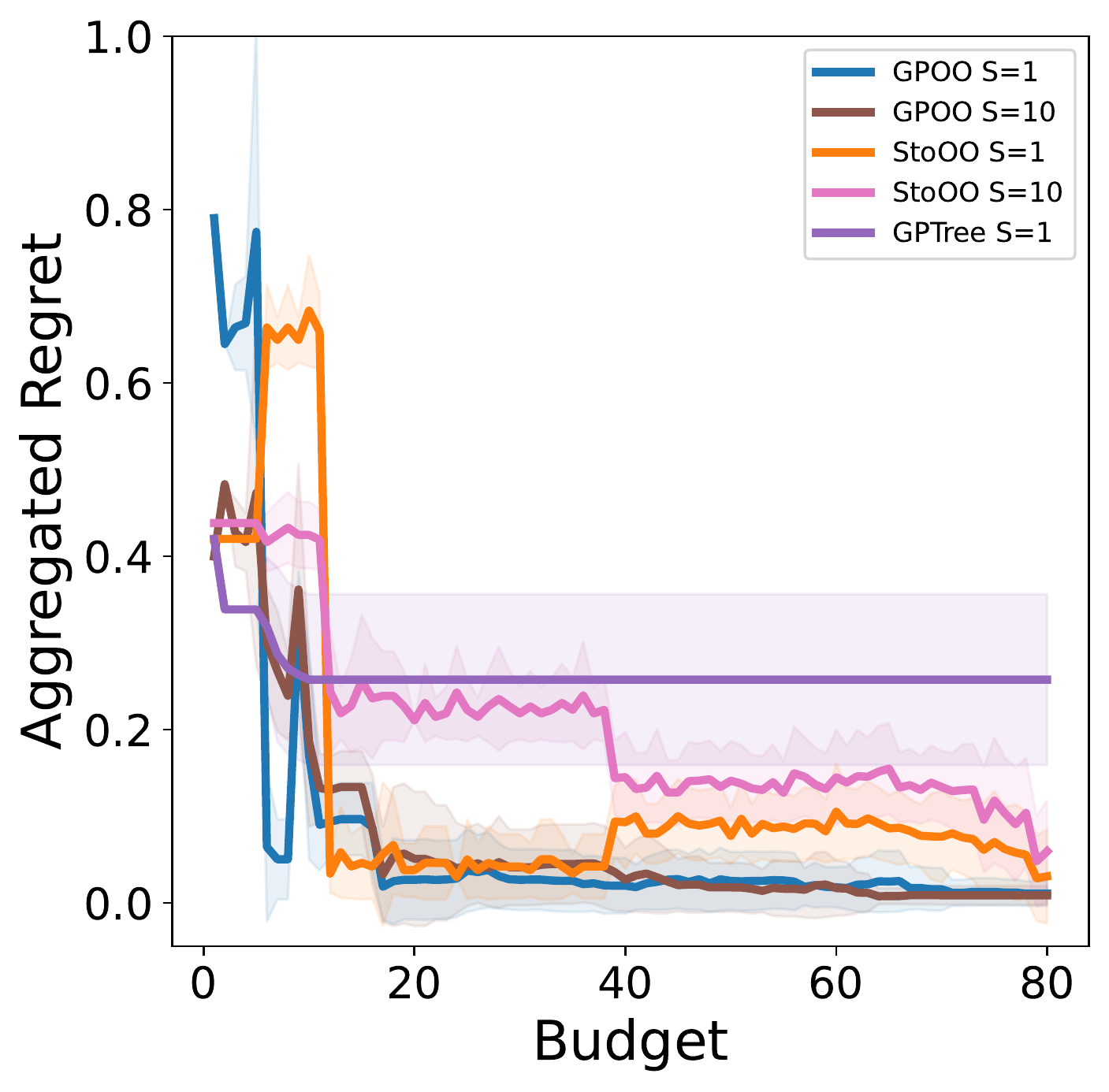}
    \includegraphics[scale=0.25]{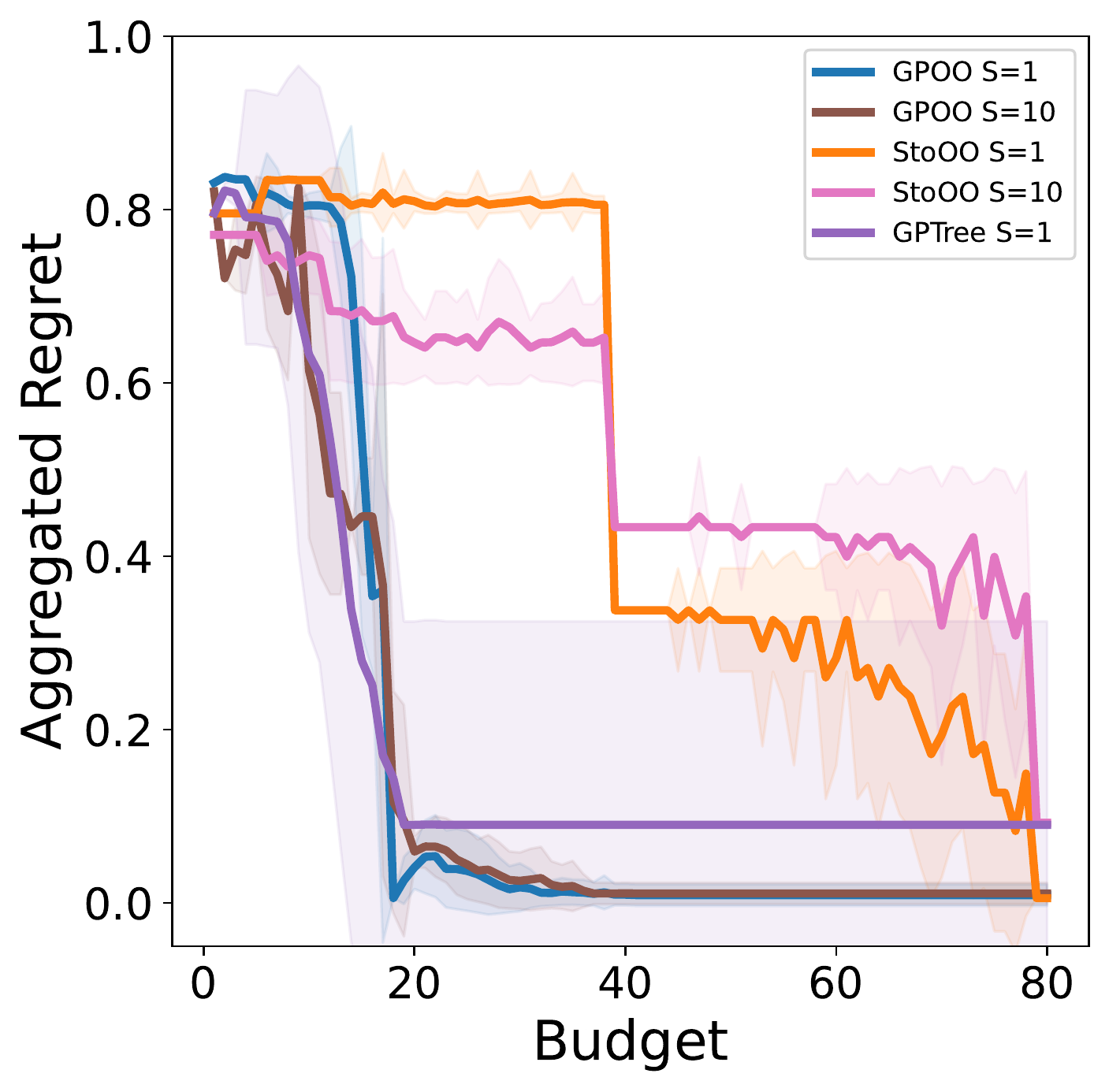}
    \caption{Reward functions $f$ (row 1) and aggregated regrets (row 2), with shaded regions indicate one standard deviation.
    We perform $30$ independent runs with a budget $N$ up to $80$.
    }
    \label{fig: regret comparison}
\end{figure}

We show regret curves of different algorithms for two simulated reward functions in Figure \ref{fig: regret comparison}.
Our decision space is chosen to be $\mathcal{X} = [0,1]$.
The reward functions are sampled from a GP posterior conditioned on hand-designed samples (listed in Appendix \ref{sec: experiment appendix}), with radial basis function (RBF) kernel having lengthscale $0.05$ and variance $0.1$.
The GP noise standard variation was set to $0.005$.
The reward noise is i.i.d. sampled zero-mean Gaussian distribution with standard deviation 0.1.

For our experiment, we consider two cases of feedback: 
\textit{single point feedback} ($S$ = 1), where the reward is sampled from the centre (representative point) of the selected cell and \textit{average feedback} ($S$ = 10), where the reward is the average of samples from the centre of each sub-cell, which are obtained by splitting the cell into intervals of equal size. 
Following Corollary~\ref{coro: Regret bound for exponential diameters}, we choose $\delta(h) = c 2^{-h}$, where $c$ is chosen via cross-validation. 
The algorithms are evaluated by the aggregated regret in Definition~\ref{defi: aggregated regret} ($S = 1$ for single point feedback, $S = 10$ for average feedback).

The related algorithms we compare with include: 
(i) StoOO \cite{munos2014treeBandits}: the error probability needed for StoOO is chosen to be $0.1$. 
(ii) AVE-StoOO: We extend the StoOO algorithm to the case where the rewards are aggregated feedback in Appendix \ref{sec: AVE-StoOO}.
(iii) GPTree \cite{shekhar2018GPBanditDiscre}.
We discuss the related algorithms in \S~\ref{sec:related_work}.

The first reward function (first row) is designed to show the performance of algorithms when there are several relatively similar local maximums. 
To find the global optimal region, the algorithm needs to predict the function values of high-value cells in high precision quickly.
The second reward function is designed to show the case where the reward function is periodic-like. 
To achieve low regret, the algorithm needs to avoid disruptions by the patterns hidden in rewards, e.g. avoid consistently sampling points with similar function values. 
GPOO (for both single and average feedback cases) has the best performance for the two reward functions. 
The aggregated regret convergences quickly and remains stable. 
GPTree tends to stuck in local maximum (in a finite budget) since their parameter $\beta_N$, balancing the posterior mean and standard deviation, only depends on the budget $N$ and does not increase over different rounds, while in our algorithm $\beta_t$ increases like $\mathcal{O}(\log t)$.
As expected, StoOO-based algorithms converge more slowly, since they only use empirical means instead of GP predictions and they are designed with a more broad family of reward functions in mind.
We also show an example where aggregated feedback can be beneficial for recommendations in Figure~\ref{fig: gpoo center ave tree}, 
which is deferred to Appendix \ref{sec: experiment appendix} due to the page limit.


\section{Related Work}
\label{sec:related_work}

We mainly review the related work in two aspects: GP- elated bandits and bandits work considering the aggregated feedback.
Refer to \citet{lattimore2018bandit} for a comprehensive review of bandits literature.

\paragraph{Gaussian Process Bandits}
By assuming the unknown reward is sampled from a GP,
bandit algorithms can be applied to black-box function optimisation.
\citet{srinivas2009_GPUCB} studied \emph{regret minimisation} under single arm feedback, 
where arms are recommended sequentially under a Upper Confidence Bound (UCB) policy.
They covered both the finite arm space and continuous arm space.
GP bandit is also studied and applied in Bayesian optimisation in a simple regret setting ~\citep{shahriari_taking_2016}.

As far as we know, no current GP bandit works address aggregated feedback. 
\citet{accabi2018GPCUCB} studied the \emph{semi-bandit} setting where both the individual labels of arms and the aggregated feedback in the selected subset are observed in each round, while we consider a harder setting where only the aggregated feedback is observed. 
For non-aggregated feedback, 
the most related work to ours is \citet{shekhar2018GPBanditDiscre}, 
which extends the StoOO algorithm to the case where the $f$ is sampled from unknown GP with theoretical analysis.
They did not provide empirical evaluations on the proposed algorithm and their algorithm highly depends on large amount of theoretical-based parameters. We compared with their algorithm in Section \ref{sec: experiments}.

\subsubsection{Aggregated Feedback Related Settings}

The ``aggregated feedback'' is first used in \citet{pike-burke_bandits_2018}, in which they studied bandits with 
\emph{Delayed, Aggregated Anonymous Feedback}.
At each round, the agent observes a delayed, sum of previously generated rewards. 
Different from our setting, they considered finite, independent arms setting and cumulative (pseudo-)regret minimisation problem.

There are two types of bandits problems that are related to the aggregated feedback setting.
One related setting is \emph{full-Bandits} \cite{rejwan2020top, du2021combinatorial}
which is studied under \emph{combinatorial bandits}~\citep{chen_combinatorial_2013} with finite arm space,
where only the aggregated feedback over a combinatorial set is observed.
No work has addresses GP bandits with full-bandit feedback. 
Our setting is different from full-bandits since we consider a continuous arm space and the objective is to identify local areas for function optimisation.
Another related setting is \emph{slate bandits}~\citep{dimakopoulou2019marginal}, where a slate has fixed number of positions named \emph{slots}.
The slate-level reward can be an aggregation over slot-level rewards. 
However, the slate bandits setting assumes slate-slot two levels rewards and each slate has fixed choices of slots, which is significantly different from our setting. 


\section{Conclusion and future work}
We introduced a novel setting for continuum-armed bandits where only the aggregated feedback can be observed. 
This is motivated by applications where aggregated reward is the only feedback or precise reward is expensive to access. 
We proposed Gaussian Process Optimistic Optimisation (GPOO) in Algorithm \ref{alg: GPOO} 
which adaptively searches a hierarchical partitioning of the space.
We provided an upper bound on the aggregated regret in Theorem \ref{theo: regret bound} and empirically evaluated our algorithm on simulated data in \S~\ref{sec: experiments}.

It may be possible to extend our framework to some other interesting settings. For example, instead of only observing the aggregated function value corrupted by Gaussian noise, one may consider the case where the gradient is additionally observed. This is recently studied in \citet{shekhar2021FirstOrderOracle} on GP bandits with cumulative regret bound. It would be interesting to study the simple regret (pure exploration), under the aggregated feedback setting.
Since GPs are closed under linear operators (including integrals, derivatives and Fourier transforms), one can potentially handle the case where some combination of these is observed. 


\bibliography{aaai22}

\begin{thebibliography}{22}
\providecommand{\natexlab}[1]{#1}

\bibitem[{Accabi et~al.(2018)Accabi, Trovo, Nuara, Gatti, and
  Restelli}]{accabi2018GPCUCB}
Accabi, G.~M.; Trovo, F.; Nuara, A.; Gatti, N.; and Restelli, M. 2018.
\newblock When Gaussian Processes Meet Combinatorial Bandits: GCB.
\newblock In \emph{14th European Workshop on Reinforcement Learning}, 1--11.

\bibitem[{Agrawal(1995)}]{agrawal1995continuum}
Agrawal, R. 1995.
\newblock The continuum-armed bandit problem.
\newblock \emph{SIAM journal on control and optimization}, 33(6): 1926--1951.

\bibitem[{Audibert and Bubeck(2010)}]{audibert2010best}
Audibert, J.-Y.; and Bubeck, S. 2010.
\newblock Best Arm Identification in Multi-Armed Bandits.
\newblock In \emph{Proceedings of the 23rd Annual Conference on Learning Theory
  (COLT)}.

\bibitem[{Bubeck, Munos, and Stoltz(2011)}]{bubeck_conti_2011}
Bubeck, S.; Munos, R.; and Stoltz, G. 2011.
\newblock Pure exploration in finitely-armed and continuous-armed bandits.
\newblock \emph{Theoretical Computer Science}, 412(19): 1832--1852.

\bibitem[{Chen, Wang, and Yuan(2013)}]{chen_combinatorial_2013}
Chen, W.; Wang, Y.; and Yuan, Y. 2013.
\newblock Combinatorial multi-armed bandit: {General} framework, results and
  applications.
\newblock \emph{30th International Conference on Machine Learning, ICML 2013},
  151--159.

\bibitem[{Dimakopoulou, Vlassis, and Jebara(2019)}]{dimakopoulou2019marginal}
Dimakopoulou, M.; Vlassis, N.; and Jebara, T. 2019.
\newblock Marginal Posterior Sampling for Slate Bandits.
\newblock In \emph{IJCAI}, 2223--2229.

\bibitem[{Du, Kuroki, and Chen(2021)}]{du2021combinatorial}
Du, Y.; Kuroki, Y.; and Chen, W. 2021.
\newblock Combinatorial pure exploration with full-bandit or partial linear
  feedback.
\newblock In \emph{Proceedings of the AAAI Conference on Artificial
  Intelligence}, volume~35, 7262--7270.

\bibitem[{Ghosal and Roy(2006)}]{ghosal2006posterior}
Ghosal, S.; and Roy, A. 2006.
\newblock Posterior consistency of Gaussian process prior for nonparametric
  binary regression.
\newblock \emph{The Annals of Statistics}, 34(5): 2413--2429.

\bibitem[{Golub and van Loan(2013)}]{golub13}
Golub, G.~H.; and van Loan, C.~F. 2013.
\newblock \emph{Matrix Computations}.
\newblock JHU Press, fourth edition.
\newblock ISBN 1421407949 9781421407944.

\bibitem[{Jansky(1933)}]{jansky1933electrical}
Jansky, K.~G. 1933.
\newblock Electrical disturbances apparently of extraterrestrial origin.
\newblock \emph{Proceedings of the Institute of Radio Engineers}, 21(10):
  1387--1398.

\bibitem[{Lattimore and Szepesvári(2020)}]{lattimore2018bandit}
Lattimore, T.; and Szepesvári, C. 2020.
\newblock \emph{Bandit Algorithms}.
\newblock Cambridge University Press.

\bibitem[{Munos(2014)}]{munos2014treeBandits}
Munos, R. 2014.
\newblock From bandits to Monte-Carlo Tree Search: The optimistic principle
  applied to optimization and planning.

\bibitem[{Pike-Burke et~al.(2018)Pike-Burke, Agrawal, Szepesvari, and
  Grunewalder}]{pike-burke_bandits_2018}
Pike-Burke, C.; Agrawal, S.; Szepesvari, C.; and Grunewalder, S. 2018.
\newblock Bandits with delayed, aggregated anonymous feedback.
\newblock In \emph{International Conference on Machine Learning}, 4105--4113.
  PMLR.

\bibitem[{Rasmussen and Williams(2006)}]{GPML2006}
Rasmussen, C.; and Williams, C. 2006.
\newblock \emph{Gaussian Process for Machine Learning.}
\newblock MIT Press.

\bibitem[{Rejwan and Mansour(2020)}]{rejwan2020top}
Rejwan, I.; and Mansour, Y. 2020.
\newblock Top-$ k $ Combinatorial Bandits with Full-Bandit Feedback.
\newblock In \emph{Algorithmic Learning Theory}, 752--776. PMLR.

\bibitem[{Shahriari et~al.(2016)Shahriari, Swersky, Wang, Adams, and
  de~Freitas}]{shahriari_taking_2016}
Shahriari, B.; Swersky, K.; Wang, Z.; Adams, R.~P.; and de~Freitas, N. 2016.
\newblock Taking the {Human} {Out} of the {Loop}: {A} {Review} of {Bayesian}
  {Optimization}.
\newblock \emph{Proceedings of the IEEE}, 104(1): 148--175.
\newblock Conference Name: Proceedings of the IEEE.

\bibitem[{Shekhar and Javidi(2021)}]{shekhar2021FirstOrderOracle}
Shekhar, S.; and Javidi, T. 2021.
\newblock Significance of Gradient Information in Bayesian Optimization.
\newblock In \emph{International Conference on Artificial Intelligence and
  Statistics}, 2836--2844. PMLR.

\bibitem[{Shekhar, Javidi et~al.(2018)}]{shekhar2018GPBanditDiscre}
Shekhar, S.; Javidi, T.; et~al. 2018.
\newblock Gaussian process bandits with adaptive discretization.
\newblock \emph{Electronic Journal of Statistics}, 12(2): 3829--3874.

\bibitem[{Srinivas et~al.(2009)Srinivas, Krause, Kakade, and
  Seeger}]{srinivas2009_GPUCB}
Srinivas, N.; Krause, A.; Kakade, S.~M.; and Seeger, M. 2009.
\newblock Gaussian process optimization in the bandit setting: No regret and
  experimental design.
\newblock \emph{International Conference on Machine Learning}.

\bibitem[{Zhang and Ong(2021{\natexlab{a}})}]{zhang_synbio2021}
Zhang, M.; and Ong, C.~S. 2021{\natexlab{a}}.
\newblock Opportunities and Challenges in Designing Genomic Sequences.
\newblock \emph{ICML Workshop on Computational Biology}.

\bibitem[{Zhang and Ong(2021{\natexlab{b}})}]{zhang2021quantile}
Zhang, M.; and Ong, C.~S. 2021{\natexlab{b}}.
\newblock Quantile Bandits for Best Arms Identification.
\newblock In \emph{International Conference on Machine Learning}, 12513--12523.
  PMLR.

\bibitem[{Zhang et~al.(2020)Zhang, Charoenphakdee, Wu, and
  Sugiyama}]{zhang2020learning}
Zhang, Y.; Charoenphakdee, N.; Wu, Z.; and Sugiyama, M. 2020.
\newblock Learning from Aggregate Observations.
\newblock \emph{Advances in Neural Information Processing Systems}, 33.

\end{thebibliography}

\newpage
\onecolumn
\appendix
\section*{Supplementary Materials for Gaussian Process Bandits with Aggregated Feedback}

We show the full proof of our theoritical analysis in \S~\ref{sec: GPOO proofs appendix}. 
In \S~\ref{app:cholesky_update}, we illustrate how to update Cholesky decompositions with new rows and column sequentially. 
We extend StoOO algorithm \cite{munos2014treeBandits} to address aggregated feedback in \S~\ref{sec: AVE-StoOO}.
In \S~\ref{app:integral_reward}, we discuss a continuous analogue of the aggregated feedback.
Lastly, we show experimental details and additional results in \S~\ref{sec: experiment appendix}.

\section{Supplementary Theoritical Analysis}
\label{sec: GPOO proofs appendix}
In this section, we show proofs of our theoretical statements in the order they are stated in the main paper. 
For the reader's convenience, we restate our theorems in the main paper whenever needed.

\Smooth* 

\label{app:gp_tail}
\begin{proof}
  The mean value theorem says that there exists some $\bm{x} \in \mathcal{X}$ such that
  \begin{align*}
      |f(\bm{x}_1) - f(\bm{x}_2)| \leq  \sup \limits_{x \in \mathcal{X}} \left| \partial f/ \partial x_j \right|  \Vert \bm{x}_1 - \bm{x}_2 \Vert_1.
  \end{align*}
  On the other hand, by Assumption~\ref{ass:gp_smoothness2} we have that $\sup \limits_{x \in \mathcal{X}} \left| \partial f/ \partial x_j \right| \leq L$ with probability $1 - a e^{- L^2 b / 2}$. Therefore,
    \begin{align*}
      \mathbb{P}\big( |f(\bm{x}_1) - f(\bm{x}_2)| \leq  L \Vert \bm{x}_1 - \bm{x}_1 \Vert_1 \big) \geq 1 - a e^{-L^2 b / 2}. 
  \end{align*}
  
  Let $\sigma_\text{max}^2 = \sigma_{\text{max}}^2 \sup \limits_{x \in \mathcal{X}} \frac{\partial^2}{\partial x_j \partial x'_j} k(x, x') \big|_{x=x'}$. By Theorem 5 of~\citet{ghosal2006posterior}, there exists some $a$ and $c$ such that for any $L>0$,
  \begin{align*}
      \mathbb{P}\left( \sup \limits_{x \in \mathcal{X}} \left| \partial f/ \partial x_j \right|  \geq L\right) \leq a e^{-c L^2 / \sigma^2_{\text{max}}}.
  \end{align*}
  Setting $b = 2c/\sigma_{\text{max}}^2$, we obtain the desired result.
\end{proof}



\xiEvent*

\begin{proof}
  From Proposition \ref{prop: high prop smoothness}, $\mathbb{P}\{f^{*} - f\left(\bm{x}_{h_{t}^{*}, j_{t}^{*}}\right) \leq L \ell(\bm{x}_{h_{t}^{*}, j_{t}^{*}}, \bm{x}^\ast)\} \geq 1-  \exp(- \frac{L^2 b}{2})$.
  By union bound, we have 
  \begin{align*}
      & \mathbb{P}\{\forall 1 \leq t \leq N, f^{*} - f\left(\bm{x}_{h_{t}^{*}, j_{t}^{*}}\right) \leq L \ell(\bm{x}_{h_{t}^{*}, j_{t}^{*}}, \bm{x}^\ast)\}\\
      = & \mathbb{P}\{\forall 1 \leq h_t^\ast \leq h_{\max}, f^{*} - f\left(\bm{x}_{h_{t}^{*}, j_{t}^{*}}\right) \leq L \ell(\bm{x}_{h_{t}^{*}, j_{t}^{*}}, \bm{x}^\ast)\}\\
      \geq & 1-  h_{\max} \exp(- \frac{L^2 b}{2}).
  \end{align*}
  We have shown a probability bound for event $\xi_1$ depending on $h_{\max}$ and $L$.

  Since $\bm{a}^\top \mu(X_{h, i} \mid Z_{t-1}) -\overline{F}(X_{h, i})$ is Gaussian, we have the tail bound
  $$
  \mathbb{P}\left\{\left|\bm{a}^\top \mu(X_{h, i} \mid Z_{t-1}) -\overline{F}(X_{h, i})\right| > CI_t(X_{h,i} )\right\} \leq e^{-\beta_{t} / 2},
  $$
  Applying
  the union bound,
  $$
  \big|\bm{a}^\top \mu(X_{h, i} \mid Z_{t-1}) -\overline{F}(X_{h, i})\big| \leq CI_t(X_{h,i} ) \hfill \forall h, \forall0 \leq i<K^{h}
  $$
  holds with probability $\geq 1-M e^{-\beta_{t} / 2}$, where $M = \sum_{h=0}^{h_{\max}} K^h$. Choosing $M e^{-\beta_{t} / 2}=\theta / \pi_{t}$ and using the union bound for $t \in \mathbb{N}$, the statement holds. For example, we can use $\pi_{t}=\pi^{2} t^{2} / 6$.

  By the union bound, we have the probability bound for event $\xi$.
\end{proof}

\IhBound*

  \begin{proof}
    Define $\mathcal{X}_{h,i}^{\geq ave} = \{\bm{x} \in \mathcal{X}_{h,i}| f(\bm{x}) \geq \bar{F}\left(X_{h, i}\right)\}$.
    At least one point in the representative set $\mathcal{C}_{h,i}$ is in $\mathcal{X}_{h,i}^{\geq ave}$.
    From Lemma \ref{lemma：GPOO expanded nodes Ih}, we know $I_h$ is the set of nodes at depth $h$ such that the point with maximum function value in that node is within $3 \delta(h)$ of the optimal state. That is, 
    \begin{align*}
      |I_h| &= |\{(h,i)|\bar{F}\left(X_{h, j}\right) \geq f^\ast - 3 \delta(h)\}|\\
      & \leq |\{(h,i)|f(\bm{x}) \geq f^\ast - 3 \delta(h), \bm{x} \in \mathcal{X}_{h,i}^{\geq ave}\}|\\
      & = |\{(h,i)|\bm{x} \in \mathcal{X}_{3 \delta(h)} \cap \mathcal{X}_{h,i}^{\geq ave}\}|.
    \end{align*}
  
    From Assumption \ref{ass: well-shaped cells}, 
    if $|\{(h,i)|\bm{x} \in \mathcal{X}_{3 \delta(h)} \cap \mathcal{X}_{h,i}^{\geq ave}\}| > C [3 \delta(h)]^{-d^{\eta, \ell}}$, 
    there exists at least more than $C [3 \delta(h)]^{-d^{\eta, \ell}}$ disjoint $\ell-$ball of radius $\nu \delta(h)$, which contradicts the definition of $d^{\eta, \ell}$ (Definition \ref{defi: near-optimality dimension}, take $\epsilon = 3 \delta(h)$).
    Chaining these together, we have $|I_h| \leq C [3 \delta(h)]^{-d^{\eta, \ell}}$.
  \end{proof}

  
  \Ih*

\begin{proof}
  Let $(h_t, j_t)$ be the node expanded at time $t$. 
  Define $\mathcal{X}_{h,i}^{\leq ave} = \{\bm{x} \in \mathcal{X}_{h,i}| f(\bm{x}) \leq \bar{F}\left(X_{h, i}\right)\}$.
  At least one point in the representative set $\mathcal{C}_{h,i}$ is in $\mathcal{X}_{h,i}^{\leq ave}$ and we denote that point as $\bm{x}_{h,i}^{\leq ave}$, i.e. 
  $\bm{x}_{h,i}^{\leq ave} \in \mathcal{X}_{h,i}^{\leq ave} \cap \mathcal{C}_{h,i}$.
  We know from Algorithm \ref{alg: GPOO}, the b-value of $h_t, j_t$ is larger than or equal to the b-value of the node $(h_t^\ast, j_t^\ast)$ whose corresponding cells contains $\bm{x}^\ast$.
  Define $\bm{x}^{\min}_{h_t^\ast, j_t^\ast} \in \argmin \limits_{\bm{x} \in X_{h_{t}^{*}, j_{t}^{*}}}f\left(\bm{x}\right)$ and  $\bm{x}^\ast \in X_{h_{t}^{*}, j_{t}^{*}}$.
  Then under event $\xi$, 
  \begin{align*}
    &\phantom{\geq} \bar{F}\left(X_{h_t, j_t}\right) \\
    & \geq \bm{a}^\top \mu(X_{h_{t}, j_{t}} \mid Z_{t-1})- CI_t(X_{h_{t}, j_{t}}) && \text{Lemma } \ref{lemma: tail bound on event xi}\\
    &= b_{h_{t}, j_{t}}(t)- 2CI_t(X_{h_{t}, j_{t}})  - \delta\left(h_{t}\right)&& \text{By}~\eqref{equ: b-value of GPOO}\\
    &\geq b_{h_{t}^{*}, j_{t}^{*}}(t)-3 \delta\left(h_{t}\right) && \text{Alg.}~\ref{alg: GPOO}, \text{ l}7,3 \\
    &\geq \bar{F}\left(X_{h_{t}^{*}, j_{t}^{*}}\right) +\delta\left(h_{t}^{*}\right)- 3 \delta\left(h_{t}\right) && \text{By}~\eqref{equ: b-value of GPOO}, \text{ event } \xi\\ 
    & \geq f\left(\bm{x}_{h_t^\ast,j_t^\ast}^{\leq ave}\right) + \sup_{\bm{x} \in X_{h_t^\ast,j_t^\ast}} L \ell(\bm{x}_{h_t^\ast,j_t^\ast}^{\leq ave}, \bm{x}) && \text{Assumption}~\ref{ass: decreasing diameters}\\
    &\qquad - 3 \delta\left(h_{t}\right) \\ 
    & \geq f\left(\bm{x}_{h_t^\ast,j_t^\ast}^{\leq ave}\right) + L\ell(\bm{x}_{h_t^\ast,j_t^\ast}^{\leq ave}, \bm{x}^\ast) -3 \delta\left(h_{t}\right) && \bm{x}^\ast \in X_{h_t^\ast,j_t^\ast}\\ 
    & \geq f^{*} -3 \delta\left(h_{t}\right). && 
    \text{Lemma } \ref{lemma: tail bound on event xi} 
  \end{align*}
\end{proof}

  \GPVarBound*
  
  \begin{proof}
    Partition the entries of $Y_{1:t}$ into two sets and then form the vectors $\bm{y}_{X}$, being the vector of reward observations at $X$, and $\bm{y}_{X^\mathsf{c}}$ being the vector of reward observations at points other than $X$. Then by the non-negativity of mutual information we have 
      \begin{align*}
      I(\bar{F}(X); \bm{y}_{X^\mathsf{c}}| \bm{y}_{X}) &\geq 0\\
      \Rightarrow  h(\bar{F}(X)|\bm{y}_{X}) - h(\bar{F}(X)| \bm{y}_{X}, \bm{y}_{X^\mathsf{c}}) &\geq 0 \numberthis     \label{equ: entropy difference group}
    \end{align*}
    Recall for univariate Gaussian random variable $Z \sim \mathcal{N}(\mu, \omega^2)$, the differential entropy of $Z$ is $h(Z) = \log (\omega  \sqrt{2 \pi e})$.
    As described in~\eqref{equ:gp_posterior}, we have 
    $\bm{a}^\top f(X)| \bm{y}_{X}, \bm{y}_{X^\mathsf{c}} \sim \mathcal{N}\left(\bm{a}^\top \mu(X\mid Z_{1:t}), \bm{a}^\top \Sigma(X | Z_{1:t}) \bm{a} \right)$, which gives the differential entropy of the second term, i.e. $h\left( f(X)| \bm{y}_{X}, \bm{y}_{X^\mathsf{c}}\right)= \log (\sqrt{\bm{a}^\top \Sigma(X | Z_{1:t}) \bm{a}}   \sqrt{2 \pi e})$.  
  
    We now evaluate the first term. Let $B \in \mathbb{R}^{t \times t}$ be invertible and let $u, v \in \mathbb{R}^t$. Recall the Sherman-Morrison formula
    \begin{align*}
        (B+u v^\top)^{-1} = B^{-1} - \frac{B^{-1} u v^\top B^{-1}}{1 + v^\top B^{-1} u}.
    \end{align*}
    
    Let $ c = \bm{a}^\top\ k\left(X, X \right) \bm{a}$, $z = \frac{c}{\sigma^2}$, and $n = T_X(t)$.
    For the first term, 
    the posterior variance of $f(\bm{x})$ given $\bm{y}_{\bm{x}}$ is 
    \begin{align*}
      \Sigma_t(\bm{x})  & = c
      -c {[1 \dots 1]}
      \left(c 
      \begin{bmatrix}
        1 & \dots & 1\\
        \dots & \dots & \dots\\
        1 &  \dots &   1
      \end{bmatrix} + \sigma^2 I
      \right)^{-1} c {[1 \dots 1]}^{T} \\
      & = c
      - {[1 \dots 1]}
      \left( 
      \frac{\sigma^2}{c} I + {[1 \dots 1]}^\top {[1 \dots 1]}
      \right)^{-1} c {[1 \dots 1]}^{T} \\
      &= c -c {[1 \dots 1]} \left( \frac{c}{\sigma^2}I - \frac{\frac{c^2}{\sigma^4} {[1 \dots 1]}^\top {[1 \dots 1]}}{1+ \frac{c}{\sigma^2} {[1 \dots 1]} {[1 \dots 1]}^\top}  \right) {[1 \dots 1]}^\top \\
      &= c - n \frac{c^2}{\sigma^2}  + n^2 \frac{\frac{c^3}{\sigma^2}}{{\sigma^2 + n c}} \\
      &= c\left(1 - \frac{nc}{\sigma^2}(1 - \frac{nc}{\sigma^2+nc})\right) \\
      &= c\left(1 - \frac{nc}{\sigma^2} \frac{\sigma^2}{\sigma^2 + nc} \right)\\
      &= c\frac{\sigma^4}{\sigma^2(\sigma^2 + nc)} \\
      &= \frac{c \sigma^2}{\sigma^2 + nc},
    \end{align*}
    which gives the differential entropy of the first term as 
    $h(f(X)|\bm{y}_{X}) = \log \left(\frac{\sqrt{2\pi e}}{\sqrt{\frac{T_X(t)}{\sigma^{2}}+\frac{1}{\bm{a}^\top k(X, X) \bm{a}}}}\right)$. 
    Then from~\eqref{equ: entropy difference group} we have 
    \begin{align*}
      \log \left(\frac{1}{\sqrt{\frac{T_X(t)}{\sigma^{2}}+\frac{1}{\bm{a}^\top k(X, X) \bm{a}}}} \sqrt{2 \pi e}\right) - \log (\sqrt{\bm{a}^\top\Sigma_t(X)\bm{a}} \sqrt{2 \pi e}) &\geq 0 \\
      \Rightarrow \frac{\sigma}{\sqrt{T_X(t)}} &\geq \sqrt{\bm{a}^\top\Sigma_t(X)\bm{a}}.
    \end{align*}
    \end{proof}

\DrawBound*

\begin{proof}
  By Algorithm \ref{alg: GPOO} and Proposition \ref{prop:gp_variance_bound}, we have 
  \begin{align}
      \delta(h) \leq CI_t(X) \leq \beta_t^{1/2} \frac{\sigma}{\sqrt{T_{X_{h,j}}(t)}}
  \end{align}
  Thus we have 
  \begin{align}
     T_{X_{h,j}}(t) \leq \frac{\beta_t \sigma^2}{\delta^2(h)}.
  \end{align}
\end{proof}

\RegretBound*
\begin{proof}
  Let $T_{X_{h, \cdot}}(N)$ be the number of times a node at depth $h$ has been evaluated up to round $N$. 
  We consider two types of nodes in terms of whether the nodes have been expanded at round $N$. 
  According to Lemma \ref{lemma：GPOO expanded nodes Ih}, for all nodes at level $h \in [0, h_{\max}]$,
  nodes that have been expanded belong to $I_h$ under event $\xi$;
  for those who have not been expanded, we track them according to their parents (which belong to $I_{h-1}$).
  Then by Lemma \ref{lemma: upper bound of Ih GPOO} \text{ and } \ref{lemma: upper bound for number of draws average}, we have
  \begin{align*}
      N \leq& \sum_{h=0}^{h_{\max }}\left|I_{h}\right| T_{X_{h, \cdot}}(N) +(K-1) \sum_{h=1}^{h_{\max }+1}\left|I_{h-1}\right| T_{X_{h-1, \cdot}}(N)\\
      \leq& K \sum_{h = 0}^{h_{\max}} C [3 \delta(h)]^{-d^{\eta, \ell}}
      \frac{\beta_{N+1} \sigma^2}{\delta^2(h)}.
  \end{align*}
  From E.q. (\ref{equ: GPOO loss bound hn defi ave}), we know that $h_N \leq h_{\max}$, and from Algorithm \ref{alg: GPOO} and Lemma \ref{lemma：GPOO expanded nodes Ih},
  \begin{align*}
      \bar{F}(X_{N}) \geq f^\ast - 3 \delta(h_{\max})  \geq f^\ast - 3 \delta(h_N),
  \end{align*}
  where $X_{N}$ is the recommended cell at round $N$. Thus, 
  \begin{align*}
      R_N = f^\ast - \bar{F}(X_N) \leq 3 \delta(h_N).
  \end{align*}
\end{proof}

\RegretCoro*

\begin{proof}
  From Theorem \ref{theo: regret bound}, we know 
  \begin{align*}
    \frac{N}{\beta_{N}} &\leq K \sum_{h = 0}^{h_n} C [4 c \rho^h]^{-d^{\eta, \ell}} \frac{\sigma^2}{\delta^2(h)}\\
    &= K C 4^{-d^{\eta, \ell}} \sigma^2 c^{-(d^{\eta, \ell} + 2)} \frac{\rho^{-(h_n + 1) (d^{\eta, \ell} + 2)}-1}{\rho^{-(d^{\eta, \ell} + 2)} - 1}\\
    &\leq c_1 \delta(h_n)^{-(d^{\eta, \ell}+2)}.
  \end{align*}
\end{proof}

  \newpage
  \section{Updating Cholesky decompositions with new rows and columns}
  \label{app:cholesky_update}
  \subsection{Solving a linear system}
  A Cholesky decomposition may be used to efficiently solve a linear system involving a positive definite matrix. More concretely, suppose $Q \in \mathbb{R}^{t \times t}$ is a positive definite matrix. Then the solution $\bm{y} \in \mathbb{R}^{t}$ to
  \begin{align*}
      Q \bm{y} = \bm{b}
  \end{align*}
  for $\bm{b} \in \mathbb{R}^{t}$ is denoted by $Q \backslash \bm{b} = Q^{-1} \bm{b}$. Letting $L L^\top=Q$ denote the Cholesky decomposition of $Q$ (obtained at a computational cost described in the next subsection), we have that $LL^\top \bm{y} = \bm{b}$, so that under the substitution $L^\top \bm{y} = \bm{z}$, we have \begin{align*}
      L\bm{z} &= \bm{b} \\
      L^\top \bm{y} &= \bm{z}.
  \end{align*}
  By exploiting the triangular structure of $L$, this system may be solved using forward and back substitution at a cost of $\mathcal{O}(t^2)$.
  
  \subsection{Updating a Cholesky decomposition}
  Following \S~4.2.9 of~\citet{golub13}, suppose we have a Cholesky decomposition $Q=LL^\top$ and we wish to find the Cholesky decomposition of
  \begin{align*}
      \begin{pmatrix}
        Q & Q_{21}^\top \\
        Q_{21} & Q_{22}
      \end{pmatrix} = \begin{pmatrix}
        L & \bm{0} \\
        L_{21} & L_{22} 
      \end{pmatrix} \begin{pmatrix}
        L & \bm{0} \\
        L_{21} & L_{22} 
      \end{pmatrix}^\top.
  \end{align*}
  Here we are only concerned with the case $Q_{21} \in \mathbb{R}^{1 \times t}$, $Q_{22} \in \mathbb{R}$, $L_{21} \in \mathbb{R}^{1 \times t}$, $L_{22} \in \mathbb{R}$ and $\bm{0}$ is a $t\times 1$ vector containing only zeros. We have that
  \begin{align*}
      Q &= L L^\top \\
      Q_{21} &= L_{21} L^\top \\
      Q_{22} &= L_{21}L_{21}^\top + L_{22}^2,
  \end{align*}
  so that we readily observe that $L_{21}^\top$ is the solution to a linear system that may be obtained by forward and back substitution at a cost of $\mathcal{O}(t^2)$ and subsequently $L_{22}$ is a scalar that may be obtained at an additional cost of $\mathcal{O}(t)$.
  


\newpage
\section{AVE-StoOO}
\label{sec: AVE-StoOO}
We consider an extension of the StoOO case under aggregated feedback, which we call AVE-StoOO. The setting of (AVE)-StoOO differs slightly to that of GPOO. We make the following alterations to the setting considered in the main paper.
\begin{itemize}
    \item the reward signal is $r_t = \overline{F}(X_{h_t,i_t}) + \epsilon_t$ where $\epsilon_t$ is a random variable satisfying $\mathbb{E}[\epsilon_t|X_{h_t,i_t}] = 0$,
    \item The reward signal $r_t$ satisfies $r_t\in[0,1]$.
\end{itemize}
We define the $b$-values of the leaves $(h,i) \in \mathcal{L}_t$ as 
\begin{align}
  \label{equ: bvalue for ave-stooo}
  \tilde{b}_{h, i}(t) := \hat{\mu}_{h, i}(t)+\sqrt{\frac{2 \log \left(t^{2} / \theta\right)}{T_{X_{h, i}}(t)}}+\delta(h),
\end{align}
where $\hat{\mu}_{h, i}(t) := \frac{1}{T_{X_{h, i}}(t)} \sum_{s=1}^{t} r_{s} \mathbf{1}\left\{ X_s = X_{h, i}\right\}$ 
is the empirical average of the rewards received in $X_{h,i}$,
and $T_{X_{h, i}}(t) := \sum_{s=1}^{t} \mathbf{1}\left\{X_s =  X_{h, i}\right\}$ is the number of times $(h,i)$ has been selected up to round $t$.
Recall the shorthand that $X_s$ is the feature matrix corresponding to the cell $\mathcal{X}_s$ selected at round $s$ and $r_s$.

\begin{algorithm}
  \caption{AVE-StoOO}
  \label{alg: AVE-StoOO}
  \begin{algorithmic}
    \State{\textbf{Input:} natural number K (-ary tree), function $f$ to be optimised, smoothness function $\delta$, error probability $\theta > 0$, budget $N$.}
    \State{\textbf{Initialisation:} tree $\mathcal{T}_0 = \{(0,0)\}$ (corresponds to $\mathcal{X}$), 
    leaves $\mathcal{L}_0 = \mathcal{T}_0$}.
    \For{$t = 1$ to $N$}
    \State{Select any ${(h_t, i_t) \in \argmax_{(h, i) \in \mathcal{L}_{t-1}} 
    \tilde{b}_{h,i}(t)}$ according to E.q. (\ref{equ: bvalue for ave-stooo}).}
    \State{Observe reward $r_t = \overline{F}(X_{h_t, i_t}) + \epsilon_t$.}
    \If{$T_{X_{h, i}}(t) \geq \frac{2 \log \left(t^{2} / \theta\right)}{ \delta(h)^{2}}$}
      \State{Expand node $(h_t, i_t)$ (partition $\mathcal{X}_{h_t, i_t}$ into $K$ subsets) and into children nodes ${\mathcal{C}_t = \{(h_t+1, i_1), \dots, (h_t + 1, i_K)\}}$.}
      \State{$\mathcal{T}_t = \mathcal{T}_{t-1} \cup \mathcal{C}_t$. $\mathcal{L}_{t-1} = \mathcal{L}_{t-1} \backslash \{(h_t, i_t)\}; \mathcal{L}_t = \mathcal{L}_{t-1} \cup \mathcal{C}_t$.}
    \EndIf
    \EndFor
    \State{\textbf{Return} The any deepest node that has been expanded $\argmax_{(h,i) \in \mathcal{T}_N \backslash \mathcal{L}_N} h.$}
  \end{algorithmic}
\end{algorithm}

\subsubsection{Analysis of AVE-StoOO}
We analyse AVE-StoOO by similar assumptions (Assumption \ref{ass: well-shaped cells}, \ref{ass: decreasing diameters stooo}, \ref{ass: local smoothness of f stooo}) made in \citet{munos2014treeBandits}.

\begin{ass}[Local smoothness of $f$]
  \label{ass: local smoothness of f stooo}
  There exists at least one global optimiser $\bm{x}^\ast \in \mathcal{X}$ of $f$ (i.e. $f(\bm{x}^\ast) = {sup}_{\bm{x} \in \mathcal{X}} f(\bm{x})$ 
  and for all $\bm{x} \in \mathcal{X}$, $f\left(\bm{x}^{*}\right)-f(\bm{x}) \leq \ell\left(\bm{x}, \bm{x}^{*}\right)$.
\end{ass}

\begin{ass}[Decreasing diameters]
  \label{ass: decreasing diameters stooo}
  There exists a decreasing sequence $\delta(h) > 0$, s.t. for any depth $d \geq 0$ and for any cell $X_{h,i}$
  of depth $h$, we have $\sup _{\bm{x} \in X_{h, i}} \ell\left(\bm{x}_{h, i}, \bm{x}\right) \leq \delta(h)$.
\end{ass}

For any $\theta \in (0, 0.5)$, define the following event
\begin{align}
  \begin{aligned}
    \tilde{\xi} := \left\{\forall h \geq 0, \forall 0 \leq i \leq K^{h}-1, \forall 1 \leq t \leq N\right.
    \left.\left|\hat{\mu}_{h, j}(t)-\bar{F}\left(X_{h, j}\right)\right| \leq \sqrt{\frac{2 \log \left(t^{2} / \theta\right)}{T_{X_{h, j}}(t)}}\right\}
    \end{aligned}
\end{align}

\begin{defi}[Martingale]
  A (discrete-time) martingale (w.r.t $Y_1, Y_2, Y_3, \dots$) is a discrete-time stochastic process 
  (i.e. a sequence of random variables) $X_1, X_2, X_3, \dots$ that satisfies for any time $n$,
  \begin{align*}
    \mathbb{E}[|X_n|] < \infty \quad \quad 
    \mathbb{E}[X_{n+1} | Y_1, \dots, Y_n] = X_n.
  \end{align*}
  
\end{defi}

\begin{theo}[Azuma's inequality]
  \label{theo: azuma}

  Let $\{X_0, X_1, \dots\}$ be a (super-)martingale w.r.t filtration $\{\mathcal{F}_0, \mathcal{F}_1, \dots\}$.
  Assume there are predictable processes $\{A_0, A_1, \dots\}$ and $B_0, B_1, \dots$ w.r.t $\{\mathcal{F}_0, \mathcal{F}_1, \dots\}$, 
  i.e. for all $t$, $A_t, B_t$ are $\mathcal{F}_{t-1}-measurable$, and constants $0 < c_1, c_2, \dots < \infty$ s.t.
    $A_{t} \leq X_{t}-X_{t-1} \leq B_{t} \quad \text { and } \quad B_{t}-A_{t} \leq c_{t}$
  almost surely. Then for all $\epsilon > 0$,
  \begin{align*}
    \mathrm{P}\left(X_{n}-X_{0} \geq \epsilon\right) \leq \exp \left(-\frac{2 \epsilon^{2}}{\sum_{t=1}^{n} c_{t}^{2}}\right).
  \end{align*}
  A submartingale is a supermartingale with signs reversed.
  If both hold, then by union bound we have a two-sided bound
  \begin{align*}
    \mathrm{P}\left(\left|X_{n}-X_{0}\right| \geq \epsilon\right) \leq 2 \exp \left(-\frac{2 \epsilon^{2}}{\sum_{t=1}^{n} c_{t}^{2}}\right).
  \end{align*} 

\end{theo}

\begin{lemma}
  For any $\theta \in (0, 0.5)$, for event $\tilde{\xi}$ defined in E.q. \ref{equ: bvalue for ave-stooo}, we have
  $\mathbb{P}(\tilde{\xi}) \geq 1 - 2\theta$.
\end{lemma}

\begin{proof}
The result follows from Azuma's inequality and a union bound.

Let $m \leq N$ be the number of nodes has been expanded up to round $t$.
For $1 \leq i \leq N$, denote $t_i$ as the time step when the $i^{th}$ node is expanded,
and $(h_{t_i}, j_{t_i})$ as the corresponding node. 
We further denote $\tau_i^s$ as time when the node $(h_{t_i}, j_{t_i})$ has been selected for the $s^{th}$ time,
and the reward obtained at that time as $r_{\tau_i^s}$.
We re-state event $\tilde{\xi}$ using the above notation, 
\begin{align}
    \tilde{\xi}=\left\{\forall 1 \leq i \leq m, \forall 1 \leq u \leq T_{{h}_{t_i}, {j}_{t_i}}(t) \quad
    \left|\frac{1}{u} \sum_{s=1}^{u} {r}_{\tau_{i}^{s}}-\bar{F}\left(X_{{h}_{t_i}, {j}_{t_i}}\right)\right| \leq \sqrt{\frac{2 \log \left(t^{2} / \theta\right)}{u}}\right\}
\end{align}

Note $\frac{1}{u} \sum_{s=1}^{u} {r}_{\tau_{i}^{s}}-\bar{F}\left(X_{{h}_{t_i}, {j}_{t_i}}\right) = \frac{1}{u} \sum_{s=1}^{u} \left({r}_{\tau_{i}^{s}}-\bar{F}\left(X_{{h}_{t_i}, {j}_{t_i}}\right)\right)$.
We have $\mathbb{E}[{r}_{\tau_{i}^{s}}|X_{{h}_{t_i}, {j}_{t_i}}] = \bar{F}\left(X_{{h}_{t_i}, {j}_{t_i}}\right)$. Let $M_u := \sum_{s=1}^{u} \left({r}_{\tau_{i}^{s}} - \bar{F}\left(X_{{h}_{t_i}, {j}_{t_i}}\right)\right)$ and note that $\{M_u: u = 0, \dots, T_{{h}_{t_i}, {j}_{t_i}}(n)\}$ is a Martingale w.r.t. the filtration generated by the samples collected at $X_{{h}_{t_i}, {j}_{t_i}}$, where $M_0 = 0$.
Given the assumption that $r_t$ is bounded between $0$ and $1$, 
we have $M_u - M_{u-1} = r_{\tau_i^u} - \bar{F}\left(X_{{h}_{t_i}, {j}_{t_i}}\right)$ for $u \geq 1$. That is we have $-1\leq M_u - M_{u-1} \leq 1$, and we make take the difference bound term in Theorem \ref{theo: azuma} to satisfy $c_u \leq 2$. Then Azuma's inequality implies that 
\begin{align*}
  & \mathbb{P}\left(\left|\frac{1}{u} \sum_{s=1}^{u} {r}_{\tau_{i}^{s}}-\bar{F}\left(X_{{h}_{t_i}, {j}_{t_i}}\right)\right|
   \geq  \sqrt{\frac{2 \log \left(t^{2} / \theta\right)}{u}}\right) \\
   = & \mathbb{P}\left(\left|\sum_{s=1}^{u} \left({r}_{\tau_{i}^{s}}-\bar{F}\left(X_{{h}_{t_i}, {j}_{t_i}}\right) \right)\right|
   \geq  u \sqrt{\frac{2 \log \left(t^{2} / \theta\right)}{u}}\right) \\
   \leq & 2 \exp\left(\frac{-4 \left( u \sqrt{\frac{\log \left(t^{2} / \theta\right)}{u}}\right)^2}{4u}\right)\\
    = & 2 \exp(\log(\theta/t^2))\\
   \leq & 2 \theta/t^2.
\end{align*}
Take a union bound over the number of samples $u \leq t$ and the number of expanded nodes $m \leq t$, we deduce the result. 
 
\end{proof}

Define the set of nodes AVE-StoOO expanded at depth $h$ as 
\begin{align}
  {\tilde{I}}_h:= \{(h,i)| \bar{F}\left(X_{h, i}\right) + 3 \delta(h) \geq f^\ast\}.
\end{align}

\begin{lemma}
  \label{lemma: AVE-StoOO only expands I}
  On the event $\tilde{\xi}$, AVE-StoOO only expands nodes that belong to the set $\tilde{I}: \cup_{h \geq 0} {\tilde{I}}_h$.
\end{lemma}

\begin{proof}
  Let $(h_t, j_t)$ be the node expanded at time $t$. 
  We know from Algorithm \ref{alg: AVE-StoOO}, the b-value of $h_t, j_t$ is large than or equal to the b-value of the cell $h_t^\ast, j_t^\ast$ containing $\bm{x}^\ast$.
  Define $\mathcal{X}_{h,i}^{\leq ave} = \{\bm{x} \in \mathcal{X}_{h,i}| f(\bm{x}) \leq \bar{F}\left(X_{h, i}\right)\}$.
  At least one point in the representative set $\mathcal{C}_{h,i}$ is in $\mathcal{X}_{h,i}^{\leq ave}$ and we denote that point as $\bm{x}_{h,i}^{\leq ave}$, i.e. 
  $\bm{x}_{h,i}^{\leq ave} \in \mathcal{X}_{h,i}^{\leq ave} \cap \mathcal{C}_{h,i}$.
  Then under event $\tilde{\xi}$,
  \begin{align*}
    \bar{F}\left(X_{h_t, j_t}\right) &\geq \hat{\mu}_{h_{t}, j_{t}}(t)-\sqrt{\frac{2 \log \left(t^{2} / \theta\right)}{T_{X_{h, j}}(t)}} && \text{On event } \tilde{\xi}\\
    &\geq \hat{\mu}_{h_{t}, j_{t}}(t)-\delta\left(h_{t}\right) && \text{By Algorithm } \ref{alg: AVE-StoOO}\\
    &\geq b_{h_{t}, j_{t}}(t)-3 \delta\left(h_{t}\right)&& \text{On event } \tilde{\xi}\\
    &\geq b_{h_{t}^{*}, j_{t}^{*}}(t)-3 \delta\left(h_{t}\right) && \text{By Algorithm } \ref{alg: AVE-StoOO} \\
    &\geq \bar{F}\left(X_{h_{t}^{*}, j_{t}^{*}}\right) +\delta\left(h_{t}^{*}\right)-3 \delta\left(h_{t}\right) && \text{On event } \tilde{\xi}\\ 
    &\geq  f\left(\bm{x}_{h_t^\ast,j_t^\ast}^{\leq ave}\right) +\delta\left(h_{t}^{*}\right)-3 \delta\left(h_{t}\right) && \text{By defintion of } \bm{x}_{h,i}^{\leq ave}\\ 
    &\geq f\left(\bm{x}_{h_t^\ast,j_t^\ast}^{\leq ave}\right) +\sup _{\bm{x} \in X_{h_t^\ast,j_t^\ast}} \ell\left(\bm{x}_{h_t^\ast,j_t^\ast}^{\leq ave}, \bm{x}\right)-3 \delta\left(h_{t}\right) && \text{By Assumption } \ref{ass: decreasing diameters stooo}\\ 
    &\geq f\left(\bm{x}_{h_t^\ast,j_t^\ast}^{\leq ave}\right) + \ell\left(\bm{x}_{h_t^\ast,j_t^\ast}^{\leq ave}, \bm{x}^\ast \right)-3 \delta\left(h_{t}\right) 
    && \bm{x}^\ast \in X_{h_t^\ast,j_t^\ast}\\ 
    & \geq f^{*}-3 \delta\left(h_{t}\right) && \text{By Assumption } \ref{ass: local smoothness of f stooo}
  \end{align*}
\end{proof}

\begin{lemma}
  Recall $d^{\eta, \ell}$ be the $\nu-$near-optimality, and $C$ be the corresponding constant. Then under Assumption \ref{ass: well-shaped cells},
  \begin{align*}
    |\tilde{I}_h| \leq C [3 \delta(h)]^{-d^{\eta, \ell}}.
  \end{align*}
\end{lemma}
\begin{proof}
    Define $\mathcal{X}_{h,i}^{\geq ave} = \{\bm{x} \in \mathcal{X}_{h,i}| f(\bm{x}) \geq \bar{F}\left(X_{h, i}\right)\}$.
    At least one point in the representative set $\mathcal{C}_{h,i}$ is in $\mathcal{X}_{h,i}^{\geq ave}$.
    From Lemma \ref{lemma: AVE-StoOO only expands I}, we know $I_h$ is the set of nodes at depth $h$ such that the point with maximum function value in that node is within $3 \delta(h)$ of the optimal cell. That is, 
    \begin{align*}
      |\tilde{I}_h| &= |\{(h,i)|\bar{F}\left(X_{h, j}\right) \geq f^\ast - 3 \delta(h)\}|\\
      & \leq |\{(h,i)|f(\bm{x}) \geq f^\ast - 3 \delta(h), \bm{x} \in \mathcal{X}_{h,i}^{\geq ave}\}|\\
      & = |\{(h,i)|\bm{x} \in \mathcal{X}_{3 \delta(h)} \cap \mathcal{X}_{h,i}^{\geq ave}\}|.
    \end{align*}
  
    From Assumption \ref{ass: well-shaped cells}, 
    if $|\{(h,i)|\bm{x} \in \mathcal{X}_{3 \delta(h)} \cap \mathcal{X}_{h,i}^{\geq ave}\}| > C [3 \delta(h)]^{-d^{\eta, \ell}}$, 
    there exists at least more than $C [3 \delta(h)]^{-d^{\eta, \ell}}$ disjoint $\ell-$ball of radius $\nu \delta(h)$, which contradicts the definition of $d^{\eta, \ell}$ (Definition \ref{defi: near-optimality dimension}, take $\epsilon = 3 \delta(h)$).
    Chaining these together, we have $|\tilde{I}_h| \leq C [3 \delta(h)]^{-d^{\eta, \ell}}$.
\end{proof}

\begin{theo}
  Let us define $h_N$ be the smallest integer $h$ s.t. 
  \begin{align}
    \label{equ: stooo hN}
    2 C K 3^{-d^{\eta, \ell}} \sum_{l=0}^{h} \delta(l)^{-(d^{\eta, \ell}+2)} \geq \frac{N}{\log \left(N^{2} / \theta\right)}.
  \end{align}
  Then with probability $1-2\theta$, the loss of AVE-StoOO is bounded as $R_N \leq 3 \delta(h_N)$.
\end{theo}

\begin{proof}
  Let $X_{N}$ is the recommended cell at round $N$. Then following~\citet{munos2014treeBandits},
  \begin{align*}
      N \leq& \sum_{h=0}^{h_{\max }} \sum_{j=0}^{K^{h}-1} T_{X_{h, j}}(N) \mathbb{I}\left\{(h, j) \in \tilde{I}_{h}\right\} 
      +\sum_{h=1}^{h_{\max }+1} \sum_{j=0}^{K^{h}-1} T_{X_{h, j}}(N) \mathbb{I}\left\{\left(h-1, j^{\prime}\right) \in \tilde{I}_{h-1}\right\} \\
      \leq & \sum_{h=0}^{h_{\max }}\left|\tilde{I}_{h}\right| \frac{2 \log \left(N^{2} / \theta\right)}{\delta(h)^2}+(K-1) \sum_{h=1}^{h_{\max }+1}\left|\tilde{I}_{h-1}\right| \frac{2 \log \left(N^{2} / \theta\right)}{2 \delta(h-1)^2} \\
      =& K \sum_{h=0}^{h_{\max }} C[3 \delta(h)]^{-d^{\eta, \ell}} \frac{2 \log \left(N^{2} / \theta\right)}{\delta(h)^2}
  \end{align*}
  From E.q. (\ref{equ: stooo hN}), we know that $h_N \leq h_{\max}$, and from Algorithm \ref{alg: AVE-StoOO} and Lemma \ref{lemma: AVE-StoOO only expands I},
  \begin{align*}
      \bar{F}(X_{N}) \geq f^\ast - 3\delta(h_{\max})  \geq f^\ast - 3 \delta(h_N),
  \end{align*}
  Thus, 
  \begin{align*}
      R_N = f^\ast - \bar{F}(X_N) \leq 3 \delta(h_N).
  \end{align*}
\end{proof}

\newpage
\section{A continuous analogue of aggregated feedback}
\label{app:integral_reward}
Here we will informally sketch an extension to a continuous analogue of~\eqref{equ: ave reward defi}. 

\paragraph{Aggregated feedback.} Recall that in our construction, each node $(h,i)$ possess two attributes: the cell $\mathcal{X}_{h,i}$ and the $S$ representative points $\mathcal{C}_{h,i}$. We may discard the representative points, and define the reward
\begin{align*}
    r_t = \int_{\mathcal{X}_{h_t,i_t}} f(\bm{x}) \, d\bm{x} + \epsilon_t.
\end{align*}
Under Assumption~\ref{ass:gp_known_cov}, such a reward is well-defined whenever the GP $f$ is integrable, i.e. the GP kernel is such that
\begin{align*}
    \int_{\mathcal{X}_{h_t,i_t}} \big| f(\bm{x})  \big| \, d\bm{x} < \infty.
\end{align*}
Such integrability is guaranteed if $\sup \limits_{\bm{x} \in \mathcal{X}_{h_t,i_t}} \mathbb{E}|f(\bm{x})| < \infty$. 

\paragraph{GP inference.} Since GPs are closed under linear operators (when the action of such operators is well-defined), the reward $r_t$ is Gaussian and the joint distribution of the rewards together with the underlying function $f$ over any set of nodes $(h,i)$ and points $\bm{x}$ is jointly Gaussian. This allows one to write $b$-values and confidence intervals for any given node.

\newpage
\section{Experiment Details}
\label{sec: experiment appendix}

\subsection{Detailed Experimental Setting}

In this section, we provide detailed experimental settings for the simulated experiments in \S~\ref{sec: experiments}.
Our experiment runs on a CPU with Python 3.7. We extended GPy (version 1.10) package for Gaussian Process Regression with aggregated feedback (average). 

We considered two reward functions, which are both the posterior mean function from a predefined Gaussian process. 
We fixed the kernel as radial basis function (RBF) kernel having lengthscale $0.05$ and variance $0.1$.
The GP noise standard variation was set to $0.005$. 
To design the shape of functions to illustrate the performance of algorithms, we predefined a list of points that the GP was conditioned on.
For the first function (left in Figure \ref{fig: regret comparison}), the predefined points are $[0.05, 0.2, 0.4, 0.65, 0.9]$ 
and the corresponding function values are $[0.85, 0.1, 0.87, 0.05, 0.98]$. 
For the second function (right in Figure \ref{fig: regret comparison}), we split the range $0$ to $0.9$ into $10$ equal-size regions, 
setting the function values of the centers of each region to $0.1$, and function values of neighbourhood points (centre plus $2/3$ of the region size) to $0.2$.
We further added a pair of points $(0.95, 0.9)$ to indicate the optimal point. 
The shape of the two functions are shown in the first row of Figure \ref{fig: regret comparison}.
To calculate the regret, we find the (empirical optimal point and function) by grid searching the arm space with size $1000.$ 
The optimal point is shown in the red point in the first row of Figure \ref{fig: regret comparison}.
We run all algorithms on these two reward functions. 

We set the arm space as $[0,1]$, number of children $K = 2$, budget $N$ ranging from $1$ to $80$. All experiments are run for $30$ independent trials. 
In the second line of \ref{fig: regret comparison}, we indicate the mean of regret curve by solid lines and one standard deviation by shaded regions. 
For GP based algorithms (GPOO, GPTree), we choose to follow our theoretical analysis and use the known GP and do not optimise the hyperparameters.
In practice, one can choose to optimise hyperparameters to potentially achieve better performance.
Our experimental goal is illustrate the regret rate and do not focus on optimising constant terms. 
In GPOO, we set $h_{\max} = 10$; $\delta(h) = 14 \times 2^{-h}$.
In GPTree, we set all parameters as listed in their theoretical analysis. Similarly, we did not optimise these parameters.

\subsection{Additional Results}
\begin{figure}[h]
  \centering
  \includegraphics[scale=0.32]{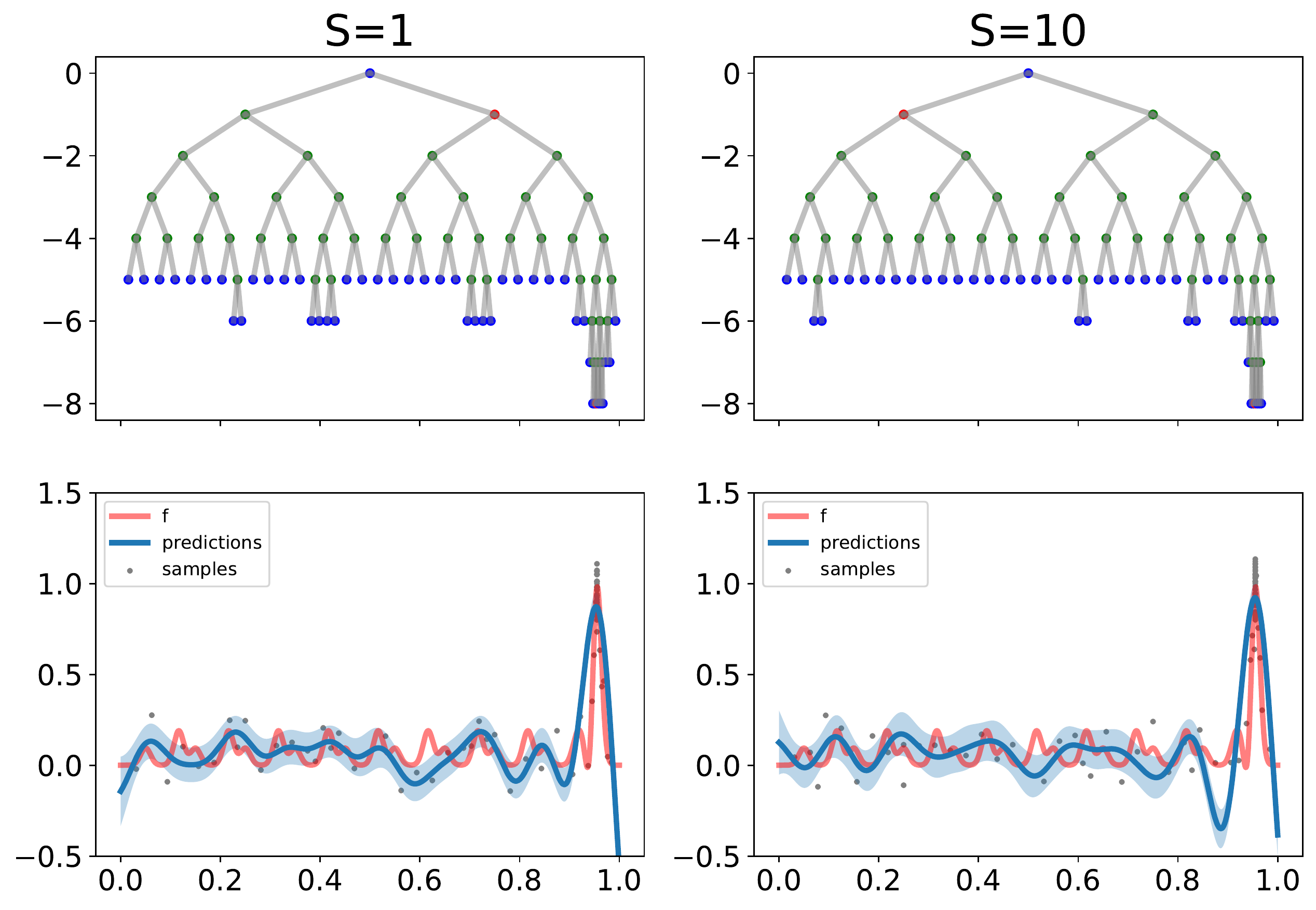}
  \includegraphics[scale=0.3]{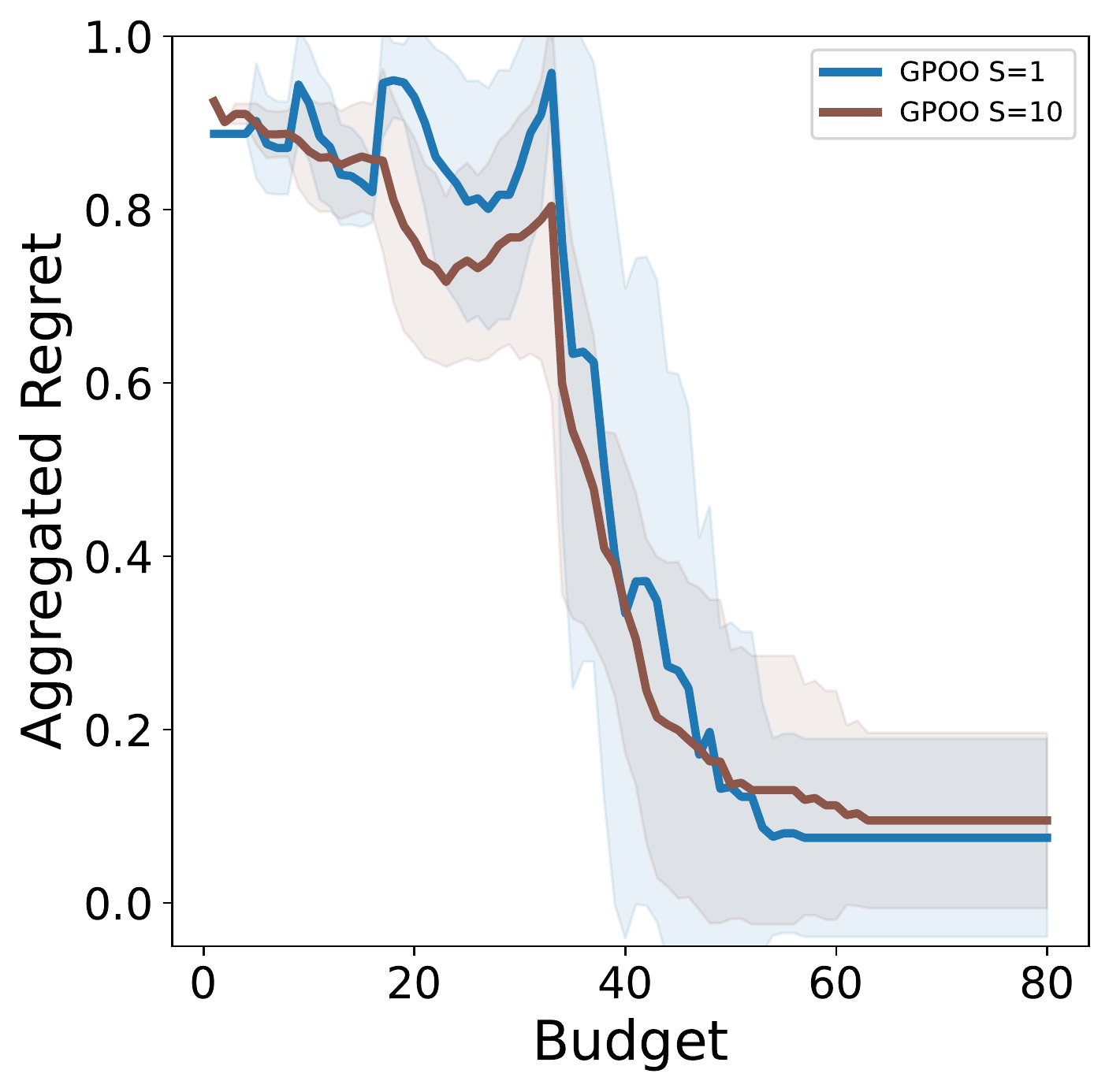}
  \caption{Example showing that aggregated feedback can be beneficial. 
  The first row shows the constructed tree with budget $N = 80$, where node colours indicate the number of times a node has been drawn (blue: 0, green: 1, red: $> 1$).
  The second row shows the GP predictions and samples at $N = 80$ (for $S=10$, the sample points are shown in centre of the selected cells).
  }
  \label{fig: gpoo center ave tree}
\end{figure}

We illustrate an example where aggregated feedback can be beneficial for recommendations in Figure~\ref{fig: gpoo center ave tree}.
The reward function was generated conditioned the same set of points from the construction of right hand function of \ref{fig: regret comparison}, plus two pairs of points $(0.94,0.1), (0.945,0.2)$. The kernel lengthscale was changed to $0.01$.
Other settings remained to be the same as above. 
This toy example shows a case where the objective function has a small amplitude, high frequency component.
The aggregated feedback can track the overall pattern of the reward functions quicker than single point feedback
The empirical regret curve  confirms that for small budget, the average feedback has smaller regret with high probability compared with centre feedback. 
Note for aggregated feedback case, the recommendation is a cell instead of a single point, so that the converged regret of $S = 10$ case is slightly larger than $S = 1$ case. 

\begin{figure}[h]
  \centering
  \includegraphics[scale=0.4]{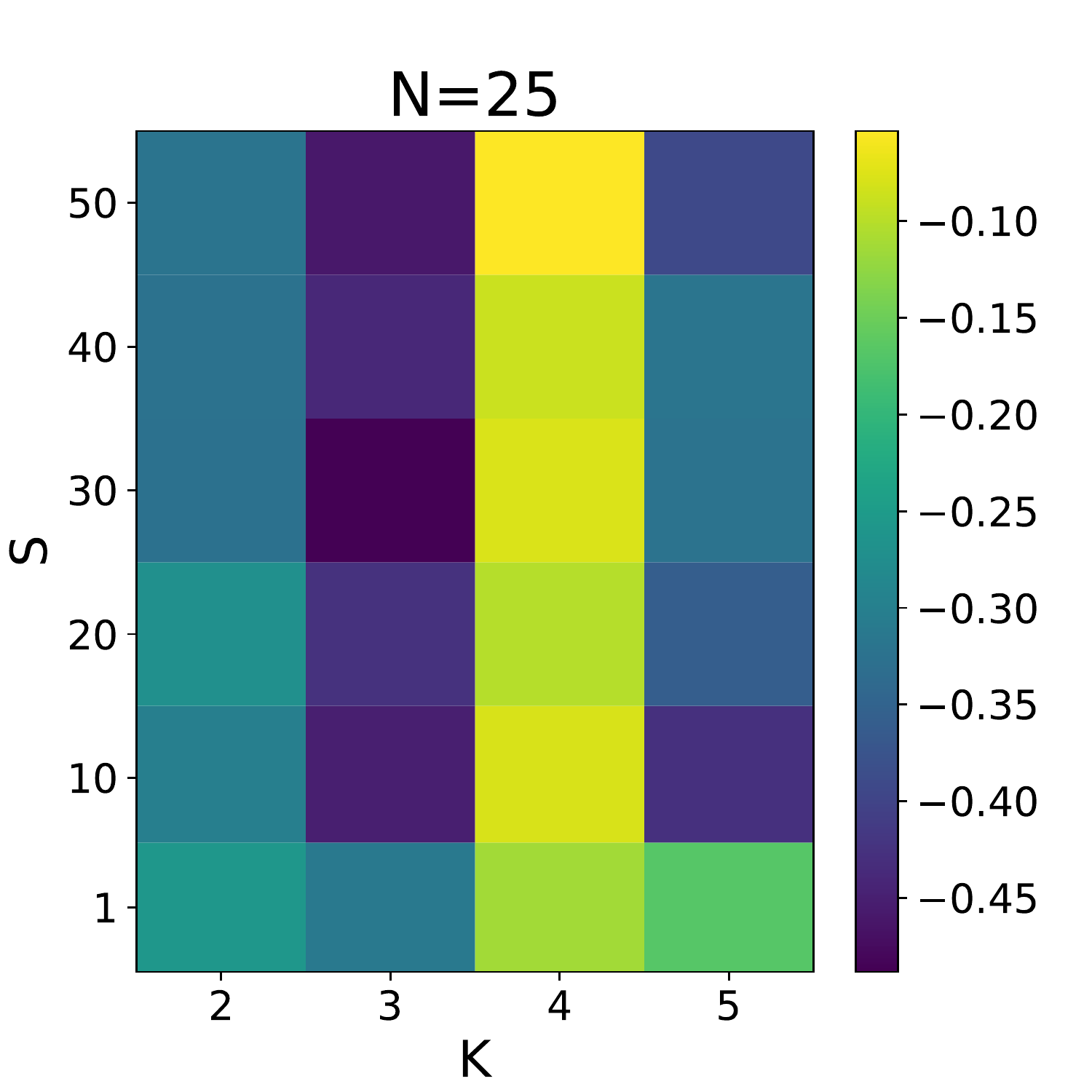}
  \includegraphics[scale=0.4]{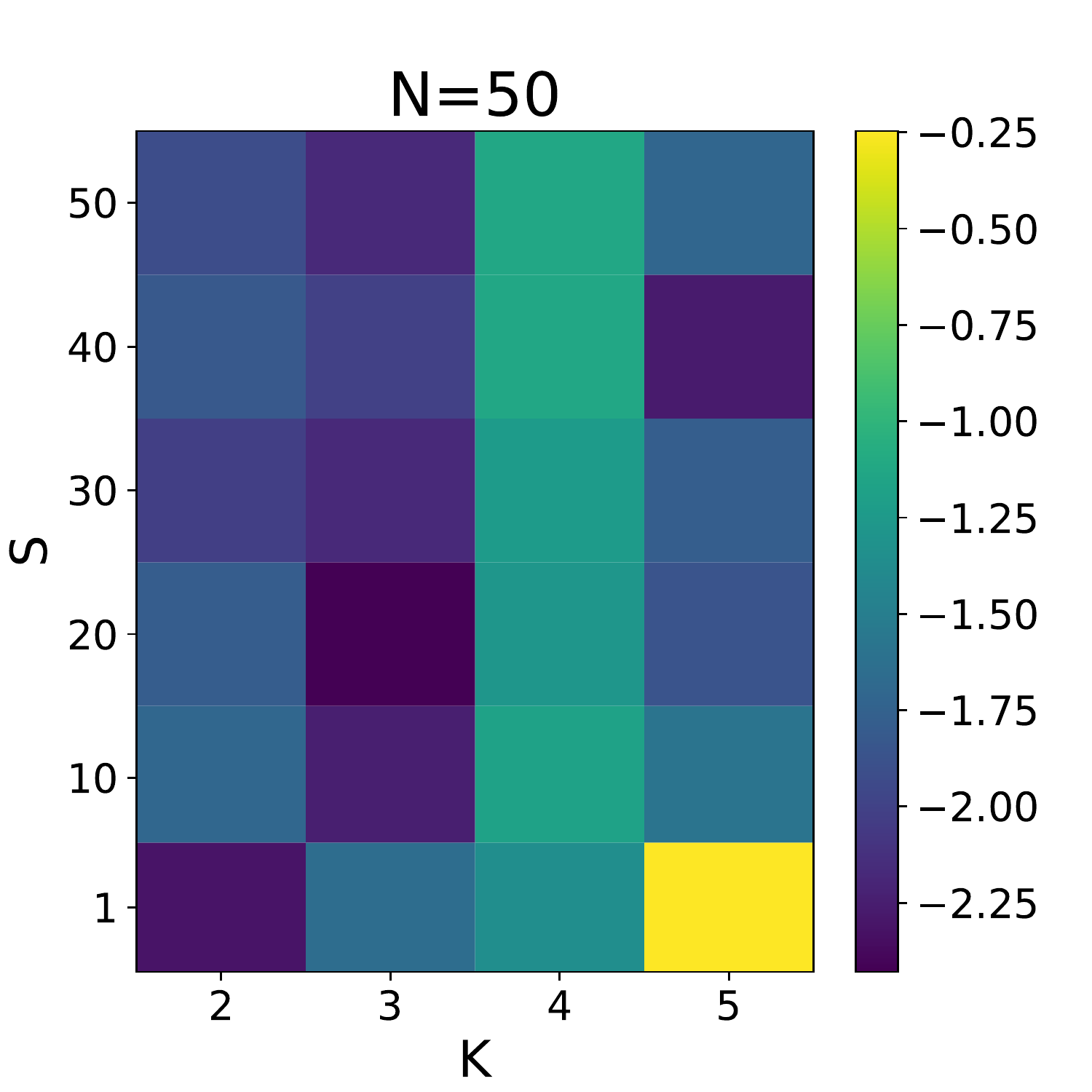}
  \caption{Logarithm Mean of Aggregated Regret for different $S$ and $K$. The colour shows the logarithm mean of aggregated regret over $30$ independent runs. 
  The smaller the colour is, the better the performance is.
  The left plot is with budget $N = 25$ and the right plot is with budget $N = 50$.
  Note the colour bar ranges are different for the two plots in order to show the regret changes in one plot clearly.}
  \label{fig: sk}
\end{figure}

We further illustrate how changing parameters ($S$ and $K$) influences the performance of GPOO algorithm. 
Based on the same reward function introduced in Figure \ref{fig: gpoo center ave tree}, we set the budget as $N = 25$ and $N = 50$, with $30$ independent runs.
We run GPOO over all combinations of the split number $S \in \{1,10,20,30,40,50\}$, and number of children $K = \{2,3,4,5\}$. 
The performance in terms of logarithm mean of aggregated regret is shown in Figure \ref{fig: sk}.
$K = 3$ appears to give the best performance in general and larger $S$ tends to provide a better choice.

\end{document}